\documentclass[sigconf]{acmart}

\usepackage{subfig}
\usepackage{multirow}
\usepackage{diagbox}
\AtBeginDocument{%
  \providecommand\BibTeX{{%
    \normalfont B\kern-0.5em{\scshape i\kern-0.25em b}\kern-0.8em\TeX}}}

\copyrightyear{2022}
\acmYear{2022}
\setcopyright{acmcopyright}\acmConference[WWW '22]{Proceedings of the ACM Web Conference 2022}{April 25--29, 2022}{Virtual Event, Lyon, France}
\acmBooktitle{Proceedings of the ACM Web Conference 2022 (WWW '22), April 25--29, 2022, Virtual Event, Lyon, France}
\acmPrice{15.00}
\acmDOI{10.1145/3485447.3511930}
\acmISBN{978-1-4503-9096-5/22/04}

\acmSubmissionID{w01fp0671}

\begin{document}

\title{Unified Question Generation with Continual Lifelong Learning}

\author{Wei Yuan}
\authornote{Equal contribution.}
\affiliation{%
  \institution{School of Information Technology and Electrical Engineering, \\The University of Queensland}
  \city{Brisbane}
  \state{QLD}
  \country{Australia}
  \postcode{4072}
}
\email{w.yuan@uq.edu.au}

\author{Hongzhi Yin}\authornotemark[1]
\affiliation{%
  \institution{School of Information Technology and Electrical Engineering, \\The University of Queensland}
  \city{Brisbane}
  \state{QLD}
  \country{Australia}
  \postcode{4072}
}
\email{h.yin1@uq.edu.au}

\author{Tieke He}\authornote{Corresponding author.}
\affiliation{%
  \institution{State Key Laboratory for Novel Software Technology, \\Nanjing University}
  \city{Nanjing}
  \country{China}}
\email{hetieke@gmail.com}

\author{Tong Chen}
\affiliation{%
  \institution{School of Information Technology and Electrical Engineering, \\The University of Queensland}
  \city{Brisbane}
  \state{QLD}
  \country{Australia}
  \postcode{4072}
}
\email{tong.chen@uq.edu.au}

\author{Qiufeng Wang}
\affiliation{%
 \institution{Department of School of Computer Science and Engineering, \\Southeast University}
 \city{Nanjing}
 \country{China}}
 \email{qfwang@seu.edu.cn}

\author{Lizhen Cui}
\affiliation{%
  \institution{School of Software, \\Shandong University}
  \city{Jinan}
  \country{China}}
  \email{clz@sdu.edu.cn}




\renewcommand{\shortauthors}{Yuan and Yin, et al.}

\begin{abstract}
  Question Generation (QG), as a challenging Natural Language Processing task, aims at generating questions based on given answers and context. 
  Existing QG methods mainly focus on building or training models for specific QG datasets.
  These works are subject to two major limitations: 
  (1) They are dedicated to specific QG formats (e.g., answer-extraction or multi-choice QG), therefore, if we want to address a new format of QG, a re-design of the QG model is required.
  (2) Optimal performance is only achieved on the dataset they were just trained on.
As a result, we have to train and keep various QG models for different QG datasets, which is resource-intensive and ungeneralizable.
  
To solve the problems, we propose a model named Unified-QG based on lifelong learning techniques, which can continually learn QG tasks across different datasets and formats.
Specifically, we first build a format-convert encoding to transform different kinds of QG formats into a unified representation.
Then, a method named \emph{STRIDER} (\emph{S}imilari\emph{T}y \emph{R}egular\emph{I}zed \emph{D}ifficult \emph{E}xample \emph{R}eplay) is built to alleviate catastrophic forgetting in continual QG learning.
Extensive experiments were conducted on $8$ QG datasets across $4$ QG formats (answer-extraction, answer-abstraction, multi-choice, and boolean QG) to demonstrate the effectiveness of our approach.
Experimental results demonstrate that our Unified-QG can effectively and continually adapt to QG tasks when datasets and formats vary.
In addition, we verify the ability of a single trained Unified-QG model in improving $8$ Question Answering (QA) systems' performance through generating synthetic QA data.
\end{abstract}




\maketitle

\section{Introduction}\label{sec_intro}
Question Generation (QG), as a dual task of question answering (QA), aims at creating grammatically and semantically precise questions according to the given text.
QG is an effective data enrichment technique for many web services and it plays a vital role in the improvement of contemporary web applications,
such as generating questions or queries from web sources to enhance QA systems~\cite{duan2017question,sun2019joint,fang2020accelerating},
guiding user-friendly interactions in dialogue systems~\cite{qi2020stay,ren2021learning}, actively enriching Frequently Asked Questions (FAQs) for web pages~\cite{krishna2019generating,liu2020asking}, and creating educational practice exercises and assessments~\cite{lelkes2021quiz}.

With advances in deep learning, deep neural network-based QG approaches have achieved remarkable progress~\cite{pan2019recent}.
However, most of the existing models are tailor-made for specific QG datasets and task formats.
For example, recent models proposed by Zhao et al.~\cite{zhao2018paragraph}, Ma et al.~\cite{ma2020improving}, and Yuan et al.~\cite{yuan2021improving} achieve state-of-the-art performance on a benchmark dataset SQuAD.
However, these models are only applicable to answer-extraction QG since they require the answers to be a span of words in context. Moreover, even for the same QG format, those models trained on one dataset are unable maintain their performance on a different one due to constrained generalizability.
Such inflexibility also exists in models built for other QG formats like multi-choice~\cite{ch2018automatic}.
In real applications, there are wide varieties of QG datasets and formats.
For example, an online shopping site needs to summarize FAQs for heterogeneous types of products with different formats, and new product types can emerge from time to time.
Since current QG models are only applicable to specific datasets and QG formats, we have to re-design, re-train, and store a large number of new models when handling different QG tasks on different datasets, which incurs serious computational and storage costs, reducing the usability and scalability of these models on large-scale web applications.
\begin{figure}[!t]
\centering
\includegraphics[width=3.3in]{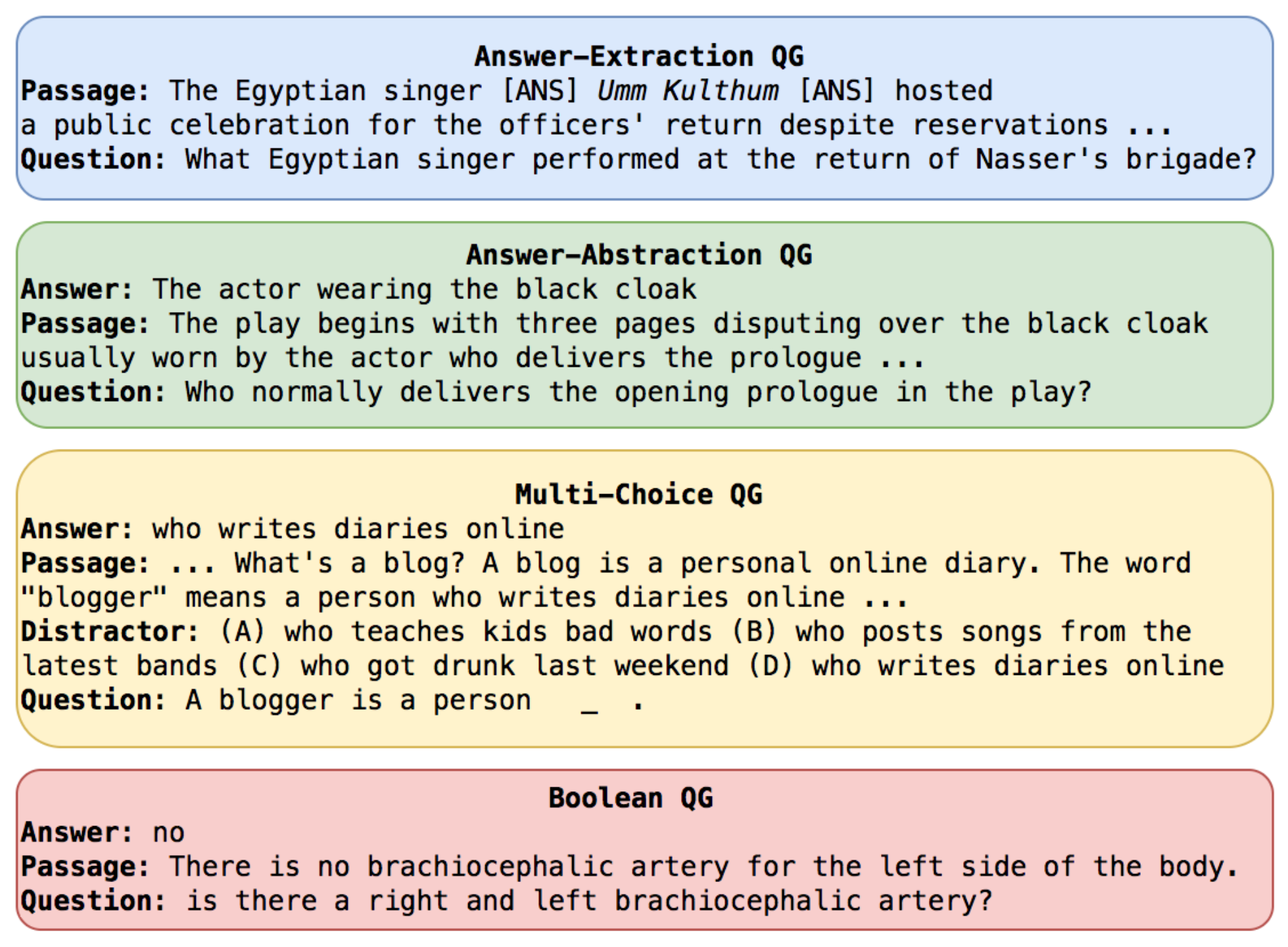}
\caption{Four formats of QG. The core components of each format are different. Samples are from SQuAD (Answer-Extraction QG), NarrativeQA (Answer-Abstraction QG), RACE (Multi-choice QG), and BoolQA (Boolean QG).}
\label{fig_qg_format}
\end{figure}

In light of this, we propose to construct a unified QG model which can generate questions across both datasets and formats.
Nevertheless, building such a unified QG model is non-trivial. 
The first challenge is that different QG formats consist of different semantic input components, which require dedicated encoding schemes from QG models to process.
Figure~\ref{fig_qg_format} shows the core components of four popular QG formats.
For answer-extraction QG, the answer is a part of the passage, while in answer-abstraction QG, the answer is not featured in the context.
The multi-choice QG contains distractors that do not exist in the other three formats, and the answer of boolean QG is simply `yes' or `no'.
Existing QG methods develop specific input interfaces for these different components respectively, lacking the ability to learn across formats.
Inspired by the recently released T5 model~\cite{raffel2019exploring}, we propose a unified QG encoding mechanism to transform each QG format into a unified representation, eliminating the boundaries brought by different input components.

Although unified QG encoding enables models to process question generation across formats, how to effectively and efficiently train a QG model across multiple datasets is still challenging.
A straightforward solution is to use multitask learning~\cite{crawshaw2020multi}, but it needs to retrain the QG model using all the historical data whenever a new dataset is available.
As a result, it is not scalable due to the linearly increasing computation and storage~\cite{biesialska2020continual} costs.
To tackle this problem, we introduce the lifelong learning concept to QG training.
Lifelong learning is more efficient than multitask learning because it does not require retraining on all historical data when adapting to new tasks (i.e., datasets in our case), which significantly reduces the computation and storage costs.
However, lifelong learning often suffers from catastrophic forgetting~\cite{mccloskey1989catastrophic,french1999catastrophic}, which means a model tends to completely and abruptly forget learned knowledge when learning new information.
To alleviate this issue, we propose a method called \emph{STRIDER} (short for \emph{S}imilari\emph{T}y \emph{R}egular\emph{I}zed \emph{D}ifficult \emph{E}xample \emph{R}eplay).
STRIDER stores a small set of prioritized samples from past datasets, and replays them every time when a new dataset is available.
This rehearsal method has been widely used in lifelong learning and achieved great success~\cite{rebuffi2017icarl,hou2019learning,chaudhry2019continual,mi2020continual}.
The main challenge of this method, especially under our QG setting, is how to select the set of prioritized data. 
In this paper, we propose a novel data selection scheme that chooses representative examples from historical data for QG based on difficulty.

Meanwhile, the size of selected samples should be small enough to reduce the storage cost, therefore, only relying on replaying prioritized data cannot fully avoid catastrophic forgetting.
To further mitigate the forgetting problem, STRIDER incorporates a parameter update regularization method, namely Elastic Weight Consolidation (EWC)~\cite{kirkpatrick2017overcoming}.
Specifically, we calculate EWC by approximating the Fisher Matrix with the chosen examples~\cite{mi2020continual} to improve computational efficiency.
Lastly, we point out that different QG datasets have different similarities.
For example, an answer-extraction QG dataset stands a good chance to be more similar to another answer-extraction dataset, rather than a multi-choice dataset.
Intuitively, for highly similar datasets, more useful knowledge from previous training steps could be preserved~\cite{ke2021adapting,ke2020continual}.
Therefore, STRIDER dynamically adjusts the weight of EWC regularization based on the similarities between the previous and current dataset, so as to adaptively control the parameter updates.

To demonstrate the effectiveness of our approach, we conduct experiments across $8$ QG datasets with $4$ commonly used QG formats mentioned in Figure~\ref{fig_qg_format}.
Experimental results show that our solution, Unified-QG, can continually learn question generation across different datasets and formats.
To sum up, our contributions are three-fold:
\begin{itemize}
  \item We propose Unified-QG to continually learn question generation across datasets and formats. To the best of our knowledge, we are the first to address the generalizability issue of QG models by subsuming QG tasks under lifelong learning.
  \item A unified QG encoding mechanism is developed to convert four popular QG formats into the text-in/text-out form, enabling QG models to learn across formats. 
        Meanwhile, a lifelong learning method named STRIDER is proposed to prevent QG models from catastrophic forgetting.
  \item Extensive experiments are conducted on $8$ QG datasets with $4$ QG formats to show our model's continual learning ability.
        The ablation studies demonstrate the importance of each component of STRIDER.
        Furthermore, we employ one \emph{single} Unified-QG to improve $8$ Question Answering (QA) models' performance on $8$ datasets respectively, which has been infeasible for traditional dedicated QG models.
\end{itemize}

\begin{figure*}[!htbp]
  \centering
  \subfloat[Traditional QG, models can only solve specific problems.]{\includegraphics[width=2.3in]{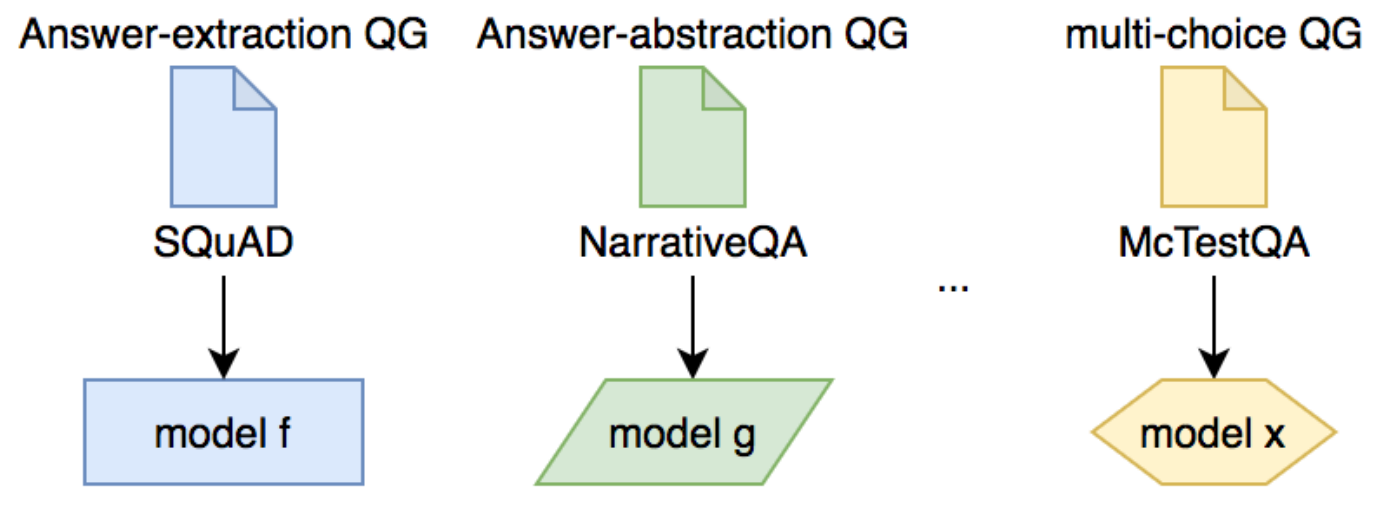}\label{fig_normal_qg}}
  \hfil
  \subfloat[QG with lifelong learning, models can continual learn to solve problem without attaching to previous datasets.]{\includegraphics[width=2.3in]{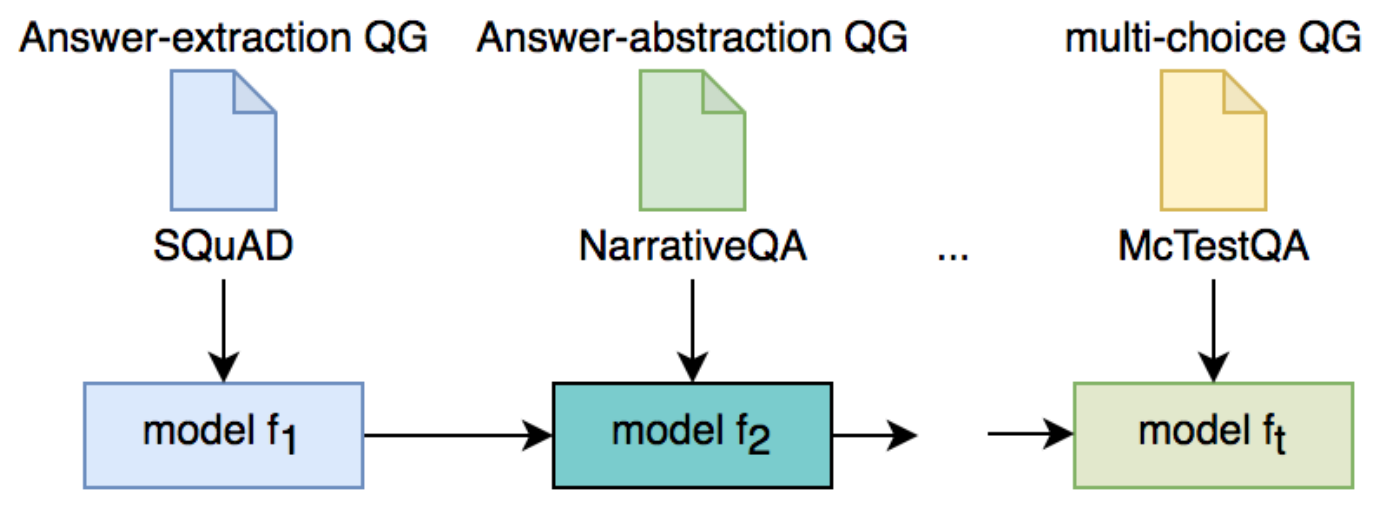}\label{fig_ll_qg}}
  \hfil
  \subfloat[QG with multitask learning, models can solve multiple tasks, but need to retrain on all previous datasets.]{\includegraphics[width=2.3in]{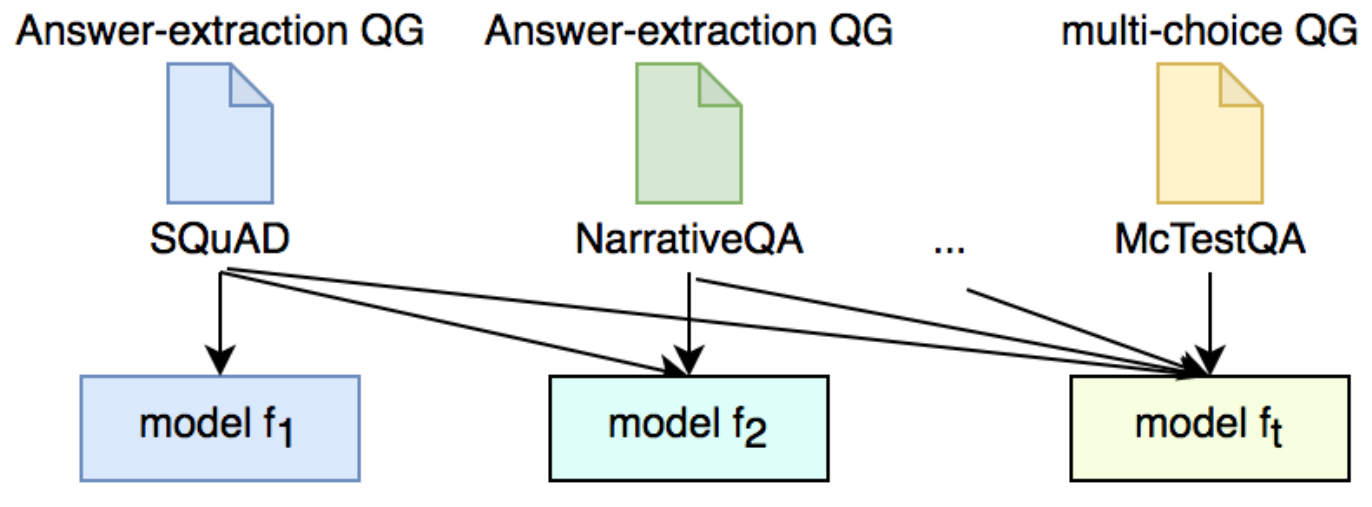}\label{fig_mul_qg}}
  \caption{The difference among traditional QG, QG with lifelong learning, and QG with multitask learning.}\label{fig_difference}
\end{figure*}

The remainder of this paper is organized as follows. 
Section 2 presents related research on question generation and lifelong learning.
Section 3 provides the problem definition and illuminates the differences among lifelong, multitask learning, meta learning, and transfer learning settings. 
Section 4 introduces Unified-QG details.
Section 5 describes the evaluation datasets and metrics, followed by results and discussions.
Section 6 concludes our work.

\section{Related Work}\label{sec:Related Work}

\subsection{Question Generation}
The emergence of large-scale datasets play a crucial role in QG systems development.
Since most of datasets are manually built from real web sources, QG systems can learn to enrich questions for web documents in a human-like way.
In principle, as a dual task of QA, any QA datasets can be used for QG~\cite{pan2019recent}.
SQuAD~\cite{rajpurkar2016squad}, MS-MARCO~\cite{bajaj2016ms} and newsQA~\cite{trischler2016newsqa} are three famous datasets used for answer-extraction QG, collected from Wikipedia, Bing search logs, and CNN news respectively.
Unlike the previous three datasets, NarrativeQA~\cite{kovcisky2018narrativeqa} does not restrict the answers to be the span of texts in the articles, therefore, it can be used as an answer-abstraction QG dataset.
Race~\cite{lai2017race}, McTest~\cite{richardson2013mctest}, OpenbookQA~\cite{mihaylov2018can}, and ARC~\cite{clark2018think} are commonly used multi-choice QG datasets.
BoolQA~\cite{clark2019boolq} is a typical boolean QG dataset, gathered from Google search engine.

Upon the variety of large-scale datasets, the QG research has made great achievements through neural network approaches.
Zhou et al.~\cite{zhou2017neural} and Du et al.~\cite{du2017learning} are the first to explore how to address the QG problem via building end-to-end neural models.
They apply a sequence-to-sequence framework with attention mechanism~\cite{bahdanau2014neural} and pointer network~\cite{gulcehre2016pointing}, which becomes a typical framework for later studies.
To alleviate the mismatch problems w.r.t. interrogative words, Sun et al.~\cite{sun2018answer} adopt an answer-aware decoding mode to generate the question words given the answer's hidden states. 
To widen the input context, Zhao et al.~\cite{zhao2018paragraph} employ gated self-attention with a maxout pointer to enhance the model's long sentence processing ability.
Yuan et al.~\cite{yuan2021improving} enrich the input information by fusing deep linguistic information based on BERT and achieve state-of-the-art performance.
However, most of these works focus on answer-extraction QG, and cannot address other QG formats, due to the assumption that the answer is a span of texts in the context.
Some works train QG models with user-written QA pairs~\cite{hu2018aspect,chali2018automatic,yu2019based,wang2019enhancing}, which are highly relevant to answer-abstraction QG.
Boolean and multi-choice QG are partially similar, where the prior one can be viewed as two-option styled multi-choice QG~\cite{ch2018automatic,lelkes2021quiz,ren2020knowledge}.
Furthermore, apart from based on text documents, many QG works attempt to generate questions with other kinds of web materials, such as knowledge graph (KG)~\cite{yu2020generating} and tables~\cite{zhu2020generating}.

Recently, applying large-scale pretrained models in QG attracts more and more researchers' interests.
Chan et al. build a recurrent BERT to output one question word at a recurrent step~\cite{chan2019recurrent,chan2019bert}, but it is time-consuming.
The generative pretrained models such as UNILM~\cite{dong2019unified}, T5~\cite{raffel2019exploring}, PEGASUS~\cite{zhang2020pegasus}, and UNILMV2~\cite{bao2020unilmv2} report the model's QG scores finetuned on SQuAD~\cite{rajpurkar2016squad} dataset, but they do not explore the idea of building a unified QG.

In this paper, we employ the pretrained model T5 as the skeleton of our Unified-QG, and conduct experiments on $8$ QG benchmarks (SQuAD, NarrativeQA, RACE, McTest, OpenbookQA, ARC-easy, ARC-hard, and BoolQA) with $4$ common formats (extraction, abstraction, multi-choice, and boolean QG).

\subsection{Lifelong Learning}
Lifelong learning (also referred to as continual learning) is a type of machine learning paradigm, which aims to continually learn across time and does not severely forget previously learned tasks~\cite{parisi2019continual}.
Since it enables models to expand their knowledge to new domains or functionalities incrementally, lifelong learning has been widely used in artificial intelligent web applications~\cite{wang2018streaming,wang2018neural,guo2019streaming,han2020graph,zhang2020gcn,he2021unsupervised}.
The main challenge in lifelong learning is that models incline to catastrophically forget existing knowledge when learning from novel tasks~\cite{thrun1995lifelong}.

Generally, approaches to alleviate catastrophic forgetting can be categorized into three families: rehearsal, regularization, and architectural methods.
Architectural methods attempt to dynamically apply modular changes or add task-specific parameters to prevent forgetting~\cite{mancini2018adding}.
However, the architectural method's parameters will dramatically increase when the number of tasks grows. 
Rehearsal methods mitigate catastrophic forgetting by retaining some training examples and replay them later.
Therefore, how to choose appropriate examples is the key challenge.
Rebuffi et al. ~\cite{rebuffi2017icarl} propose iCaRL which selects training data using Herding techniques~\cite{welling2009herding}.
Ramalho et al. ~\cite{ramalho2019adaptive} collect less confident examples and replay them periodically.
To further reduce the number of examples stored, some methods employ autoencoders~\cite{kemker2017fearnet} and GANs~\cite{shin2017continual} to generate examples by approximating previous task examples' data distribution.
Regularization methods consolidate learned knowledge via specific loss terms.
EWC~\cite{kirkpatrick2017overcoming} is a regularization term to slow down the update for parameters that are important to previous tasks.
Aljundi et al.~\cite{aljundi2018memory} measure parameters' importance based on how sensitive the output is to the changes of parameters.
Other works rectify models' parameter updates according to knowledge distillation~\cite{li2017learning,castro2018end,zhao2020maintaining,he2021automl}.

In this paper, we present STRIDER, which incorporates both rehearsal and regularization to avoid catastrophic forgetting.

\section{Problem Description}
\subsection{Question Generation}
Generally, the goal of QG is to generate a question $Q$ given a context $C$ and an answer $A$. Formally:
\begin{equation}
  \label{eq_problem_statement}
  \hat{Q} = \mathop{argmax}\limits_{Q}P(Q|C,A)  
\end{equation}
The context $C$, answer $A$, and question $Q$ are composed of a list of words.
Eq.~\ref{eq_problem_statement} is a unified formal statement of QG.
With different assumptions of $C$, $A$, and $Q$, QG can be classified into different types.
For answer-extraction QG, answer $A$ is a subspan of $C$.
While for answer-abstraction QG, answer $A$ can be any novel words.
For multi-choice QG, the context not only includes description, but also the distractors.
For boolean QG, the word in an answer is restricted to the binary ``yes'' or ``no''.
The QG dataset $D$ consists of the question-answer-context triples $D=\{(c_{i},a_{i},q_{i})\}_{i=0}^{n}$.

\subsection{Lifelong Learning of QG}\label{subsec_ll_of_qg}
As mentioned before, current QG models are dedicated to specific datasets with specific formats, and few research studies QG across datasets and formats.
In this paper, we implement such QG by using lifelong learning strategies.
Following previous lifelong learning works' setting~\cite{rebuffi2017icarl,kirkpatrick2017overcoming,mi2020continual}, we assume each QG dataset represents a QG task and the new QG data arrives task by task.
Let $f_{t}$ be the well-trained QG model in task $t$.
The new dataset $D_{t+1}$ for task $t+1$ is used to update model $f_{t}$.
After updating, the newest model $f_{t+1}$ should have good performance in all learned tasks $1:t+1$.
Different from multitask learning, lifelong learning does not require retraining the model $f_{t}$ on all the datasets $D_{1:t+1}$ when a new task $t+1$ comes in.
Figure~\ref{fig_difference} illustrates the differences among traditional QG, QG with multitask learning, and QG with lifelong learning.
Traditional QG models need to be specifically designed for each task. 
Multitask learning QG should be retrained on all previous datasets when a new task arrives,
while lifelong learning QG only needs to be continually trained on the new dataset and will perform well in all presented tasks.

It is also important to point out the setting's differences among lifelong learning, transfer learning and meta learning.
Transfer learning aims to transfer the knowledge from the source task to the target task in order to improve model performance on the target task. 
It does not consider the performance of previous tasks~\cite{peng2020few,biesialska2020continual}.
On the other hand, meta-learning focuses on learning generic knowledge from a small set of data and numerous tasks, so that it can quickly adapt to a new task.
Therefore, it enhances the learning ability without considering performance collapse for past tasks.

In the experiment section, we will discuss the superiority of our lifelong learning method for Unified-QG, compared with transfer learning and multitask learning.

\section{Unified QG with Lifelong Learning}

\subsection{Overview of Unified QG}
\begin{figure}[!h]
  \centering
  \includegraphics[width=3.3in]{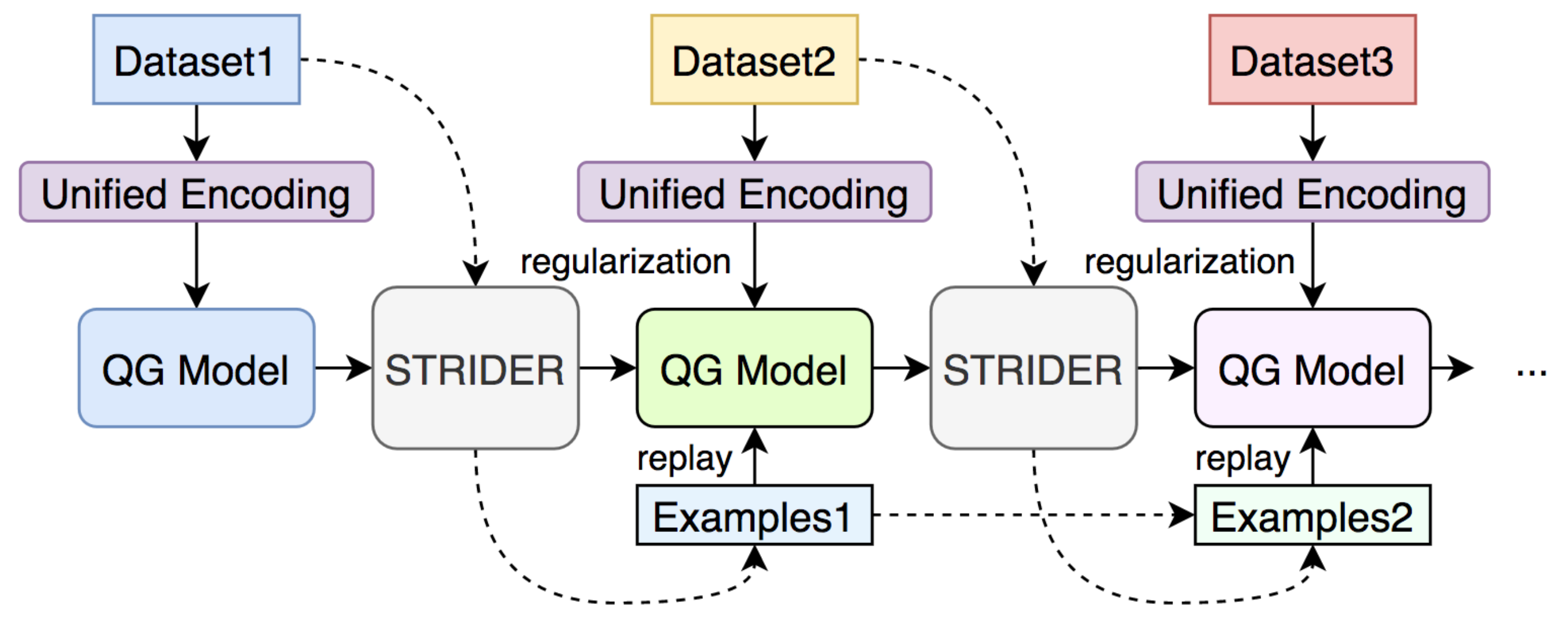}\caption{Overview of our Unified-QG approach.}
  \label{fig_overview}
\end{figure}

We first present an overview of our Unified-QG from two aspects: the model architecture and the lifelong learning method STRIDER, which are depicted in Figure~\ref{fig_overview}.
For the model architecture, we propose a unified QG encoding method to convert different QG formats into a unified form, endowing QG models with the ability to handle diversified formats.
Meanwhile, we adopt T5~\cite{raffel2019exploring} as the backbone of our QG model.
T5 is instructed to perform particular tasks by adding prefixes.
During continual training, it will identify the prefix meaning we added in the unified QG encoding.
Although the modified QG model can now produce questions across formats with unified QG encoding, it still suffers from the forgetting problem.
Hence, we propose STRIDER to alleviate this problem.
STRIDER addresses forgetting by retaining and replaying some challenging and representative examples of historical data, and elastically regularizes the model's update.

\subsection{Model Architecture}
\subsubsection{Unified QG Encoding}
As shown in Figure~\ref{fig_qg_format}, different QG formats contain different components.
Traditional QG models need specific input interfaces to handle certain formats.
In order to build a unified QG that can work across formats, we propose an encoding scheme for QG instances to unify various formats.

\begin{figure}[!h]
  \centering
  \includegraphics[width=3.3in]{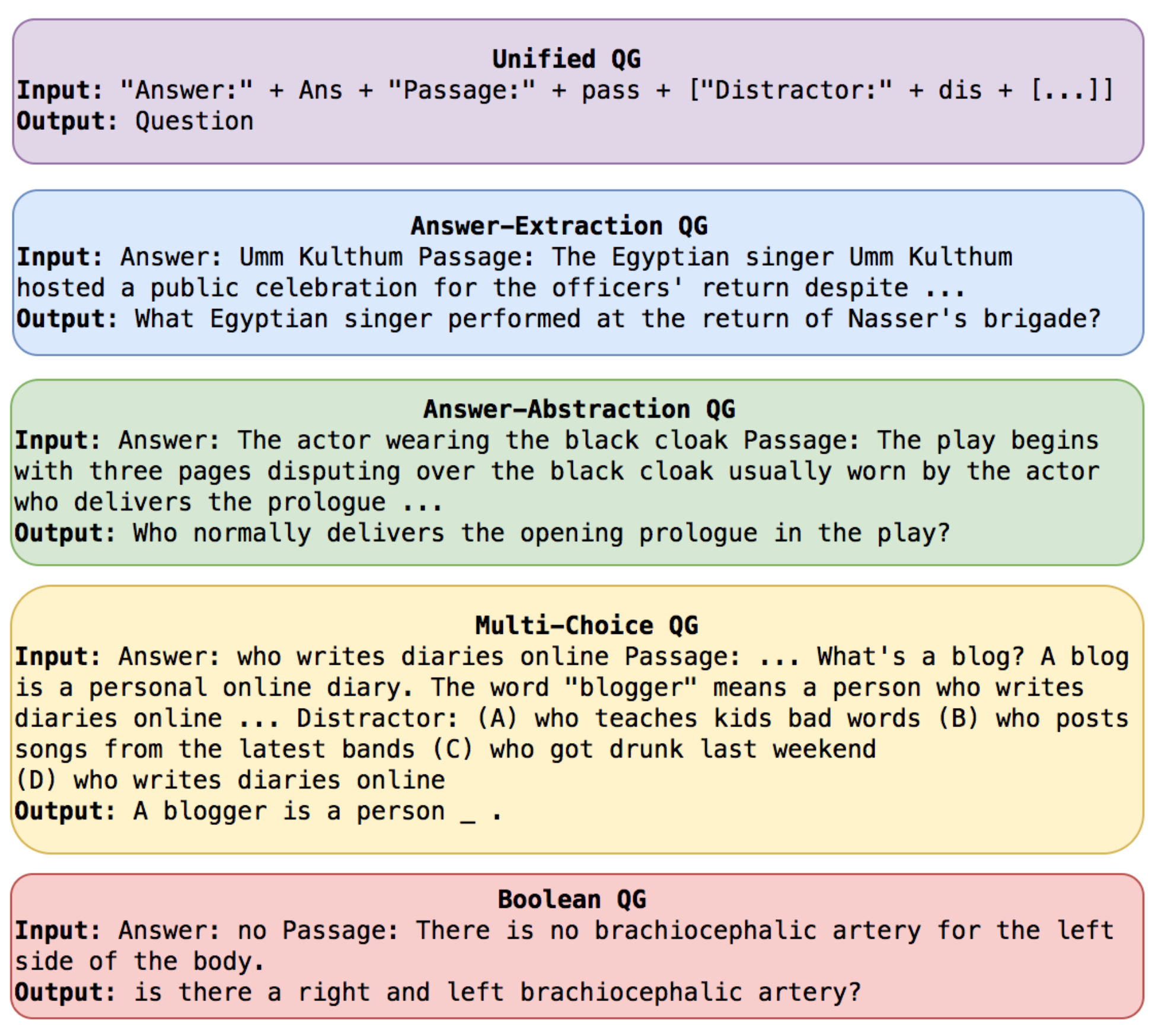}\caption{The unified QG encoding with four examples. Samples are the same as in Figure~\ref{fig_qg_format}.}
  \label{fig_unified_format}
\end{figure}

The multi-format encoding is inspired by T5 model's text-to-text encoding method.
In the unified QG encoding, all the components are concatenated as a long paragraph. 
The concatenated components have an order of answer, passage, and distractors and other components if necessary, where each component starts with a prefix.
The concatenated paragraph is then fed into our Unified-QG model and outputs the question.
Figure~\ref{fig_unified_format} shows the form of our unified QG encoding with four representation examples of different QG formats.
We emphasize that although we mainly conduct experiments with these four kinds of QG formats, our unified QG encoding can easily generalize to other QG formats.
After unified encoding, the QG dataset can be represented as $D=\{(x_{i}, y_{i})\}_{i=0}^{n}$. $x_{i}$ is the concatenated input in the Figure~\ref{fig_unified_format} and $y_{i}$ is the target question.
$x_{i}$ is then fed into the skeleton of QG model, T5, which will be introduced in the following part.

\subsubsection{T5 for QG}
We incorporate T5~\cite{raffel2019exploring} as the skeleton of our QG model, which is a variant of the encoder-decoder transformer model~\cite{vaswani2017attention} pretrained in multiple tasks. 
It allows for different input components specified by the prefixes in the input.
Its encoder is a pile of stacked transformer blocks, each of which contains two neural structures: a multi-head self-attention layer~\cite{parikh2016decomposable} and a position-wise feed-forward network:
\begin{equation}
  \begin{split}
    &x = LayerNorm(x + SelfAttn(x))\\
    &x = LayerNorm(x + FeedForward(x))
  \end{split}
\end{equation}
where $LayerNorm(\cdot)$, $SelfAttn(\cdot)$, and $FeedForward(\cdot)$ represent layer normalization~\cite{ba2016layer}, self-attention, and feed-forward network, respectively.
For simplicity, the layer normalization operation only rescales the activations and does not apply additive bias.
The residual connectors~\cite{he2016deep} are used between each two sub-layers to accelerate the speed of training.
Dropout operation~\cite{srivastava2014dropout} is adopted in all sublayers and the input and output of the entire stack.

The structure of the decoder is much similar to the encoder but it includes attention mechanism after each self-attention, in order to attend to the encoder's output.
Besides, the decoder's self-attention is causality-enabled, i.e., being only dependent on the past and present words.
The output of the decoder is then fed into a dense layer with a softmax operation to predict the question words.

\subsection{SimilariTy RegularIzed Difficult Example Replay (STRIDER)}
QG models equipped with the unified QG encoding are able to produce questions across formats.
However, it is subject to catastrophic forgetting, if only updated based on the new dataset/task.
To alleviate such forgetting, we propose a novel lifelong learning approach for QG named STRIDER. 
STRIDER is featured with similarity-based adaptive regularization and difficult data replay. 

\subsubsection{Difficult Example Replay}
To avoid catastrophically forgetting the learned knowledge from previous datasets, we propose to retain a small set of examples from the past training data and replay these examples in later training tasks.
Such replay methods have been successful in many lifelong learning works~\cite{rebuffi2017icarl,hou2019learning,chaudhry2019continual,mi2020continual}.

The effectiveness of replay methods largely depends on the quality of selected examples.
In this paper, we propose a difficult example selection method to select informative and diverse data from previous datasets.
The intuition is that, difficult training examples are more informative and helpful for improving the current model. 
Hence, with the limited replay size, we can only retain a small set of difficult historical examples for replay.

Our difficult example selection criterion on task $t$ is as follows:
\begin{equation}
  S_{t}(x_{i}^{t}, y_{i}^{t}) = \frac{CE(x_{i}^{t}, y_{i}^{t}|\theta_{t})}{\left|y_{i}^{t}\right|}
\end{equation}
where $x_{i}^{t}$ and $y_{i}^{t}$ is the $i$-th pair of training data in $D_{t}$. 
$CE(\cdot|\theta_{t})$ is the cross-entropy function calculated based on model parameter $\theta_{t}$, which can be interpreted as the model's training loss on the $i$-th sample of dataset $D_t$. 
We observe that long sentences tend to have higher accumulated loss values, therefore, 
we use the inverse of $y$'s length as a normalization factor to eliminate the impact of length on loss values. Essentially, $S_{t}(\cdot)$ quantifies the confidence of the model when performing QG for every given $x_i^t$, where a higher value indicates that the selected training sample is more challenging for the model to learn. We collect the $(x_{i}^{t}, y_{i}^{t})$ pairs that can maximize $S(x, y)$ as the difficult example set from $D_{t}$:
\begin{equation}
  \label{eq_diff_exam}
  E_{t} = \mathop{argmax}\limits_{E_{t}\subseteq D_{t} \land \left| E_{t} \right|= N} \sum_{(x,y)\in E_{t}} S_{t}(x,y)
\end{equation}
where Eq.~\ref{eq_diff_exam} aims to select $N$ examples of $(x_{t},y_{t})$ pairs that can maximize the selection criterion.

Finally, during the continual training of the unified QG model $f_{t}$ on dataset $D_{t}$, the selected example sets from previous datasets $E_{1:t-1}=\{E_{i}\}_{i=1}^{t-1}$ are integrated to augment $D_{t}$.
The training objective is to minimize the following function:
\begin{equation}
  L_{R}(\theta_{t}) = \sum\limits_{(x,y)\in D_{t} \cup E_{1:t-1}} CE(x,y|\theta_{t})
\end{equation}

\subsubsection{Similarity Regularization}
The size of retained historical examples should be as small as possible to reduce storage overhead such that $|E_{1:t}| \ll |D_{1:t}|$.
With a small $N$, the effect of replaying difficult examples on alleviating catastrophic forgetting is largely restricted due to insufficient samples for each task.
To further mitigate forgetting during continual learning, we propose a similarity-based adaptive regularization method \emph{Similarity Regularization}. 

Similarity regularization utilizes the well-established Elastic Weight Consolidation (EWC)~\cite{kirkpatrick2017overcoming} as the main regularization term.
Inspired by the idea that synaptic consolidation achieves continual learning via reducing the plasticity of synapses in human brains, EWC imposes restrictions on parameters that are important for historical tasks:
\begin{equation}
  \label{eq_ewc}
  L_{E}(\theta_{t}) = \sum\limits_{i}\lambda F_{i}(\theta_{t,i}-\theta_{t-1,i})^{2}
\end{equation}
where $F_{i}=\bigtriangledown^{2}CE(x,y|\theta_{t-1,i})$ s.t. $(x,y) \in E_{1:t-1}$ denotes the i-th element in the diagonal Fisher Information Matrix.
The Fisher Matrix describes the covariance of gradients, which is a measure for the importance of parameters.
Note that we calculate $F_{i}$ only with the example sets, without the need for storing the whole previous data.
$\lambda$ controls the influence of the EWC term.
As shown in Eq.~\ref{eq_ewc}, EWC elastically slows the changes of parameters that are important for historical tasks, and give the ``unimportant'' parameters more plasticity.

In QG, we observe that the similarities among datasets vary significantly. 
For example, datasets that belong to answer-extraction QG (e.g. SQuAD) might be more similar to answer-abstraction QG datasets (e.g. NarrativeQA) than to multi-choice QG datasets (e.g. RACE).
Naturally, the model will need less parameter updates on the dataset that is similar to previous data than on the dissimilar one.
Therefore, we propose a similarity-based adaptive weight $\lambda$ to control EWC regularization.
To approximate this similarity, we treat the current dataset and previous data as two documents and vectorize them by using TF-IDF.
TF-IDF is often used to measure the relevance between texts~\cite{ramos2003using,sugiyama2003refinement,qaiser2018text}, and is parameter-free and highly efficient to compute compared with other alternatives.
Then, we employ cosine similarity to calculate the adaptive weight $\lambda$:
\begin{equation}
  \lambda = \lambda_{ori} \frac{T(D_{t})^{\top}\cdot T(E_{1:t-1})}{\parallel T(D_{t}) \parallel \parallel T(E_{1:t-1})\parallel}
\end{equation}
where $\lambda_{ori}$ is a hyper-parameter that represents the original weight of EWC, and $T(\cdot)$ is the TF-IDF function that vectorizes a dataset.
Note that we use the selected example sets to represent the historical data instead of using all the historical data.
In general, $\lambda$ increases when the current set is similar to historical examples, thus reducing the degree of parameter updating on task $t$.

\section{Experiments}

\subsection{Datasets}
We select $8$ representative datasets that cover $4$ QG formats as our evaluation datasets, including:
\begin{itemize}
  \item Answer-extraction QG: SQuADv1.1~\cite{rajpurkar2016squad}
  \item Answer-abstraction QG: NarrativeQA~\cite{kovcisky2018narrativeqa}
  \item Multi-choice QG: RACE~\cite{lai2017race}, McTest~\cite{richardson2013mctest}, OpenbookQA~\cite{mihaylov2018can}, ARC-easy, ARC-hard~\cite{clark2018think}
  \item Boolean QG: BoolQA~\cite{clark2019boolq}
\end{itemize}
We assume datasets arrive in the following order: ``McTest$\rightarrow$SQuAD\\$\rightarrow$RACE$\rightarrow$NarrativeQA$\rightarrow$\text{Arc-easy}$\rightarrow$Arc-hard$\rightarrow$OpenbookQA\\$\rightarrow$BoolQA'', which corresponds to the exact \emph{release dates} of these datasets in the real world.
Details on dataset characteristics, statistics, and splitting strategies are in Appendix~\ref{sec:data_details}.

\subsection{Evaluation Metrics}\label{subsec_eval_metrics}
Following existing QG works~\cite{yuan2021improving,zhou2019question}, we adopt the widely used n-gram based automatic evaluation metrics, BLEU-1$\sim$4~\cite{papineni2002bleu}, METEOR~\cite{denkowski2014meteor}, and ROUGE-L~\cite{lin2004rouge} ,to evaluate our models.
BLEU measures the prediction quality based on the co-occurrence n-gram frequency between references and predictions~\cite{callison2006re}.
BLEU-1 measures the precision of unigram, while higher-order BLEU introduces a bias towards sentence fluency~\cite{song2013bleu}.
METEOR estimates candidate text's quality by calculating the harmonic mean of unigram precision and recall based on not only exact word matching, but also stemming and synonymy matching.
ROUGE-L is the Longest Common Subsequence (LCS) based ROUGE~\cite{lin2004looking}.
To some extent, it takes into account the sentence-level structure similarity between candidate text and reference and can identify the longest co-occurring in sequence n-grams.
In the following tables, we use ``B1$\sim$4'', ``M.'', and ``RL'' represent BLEU-1$\sim$4, METEOR, and ROUGE-L.

Since none of the existing work has studied QG across datasets and formats by using lifelong learning techniques, we employ two additional metrics to better evaluate the lifelong learning ability:
\begin{align}
  \mathcal{M}_{seen} &= \frac{1}{T}\sum\limits_{i=1}^{T} \mathcal{M}_{seen,i},  &\mathcal{M}_{seen,i} = \frac{1}{i}\sum\limits_{t=1}^{i} \mathcal{M}_{t,i} \label{eq_m_seen}\\
  \mathcal{M}_{first} &= \frac{1}{T}\sum\limits_{i=1}^{T} \mathcal{M}_{first,i}, &\mathcal{M}_{first,i} = \mathcal{M}_{1,i}\label{eq_m_first}
\end{align}
where $\mathcal{M}_{seen,i}$ denotes the model's average performance on all \emph{seen} tasks after the task $i$ has been learned.
$\mathcal{M}_{t,i}$ represents the model's performance on $t$-th task after learning task $i$.
$\mathcal{M}_{seen}$ is the average of the model's $\mathcal{M}_{seen,i}$ on all tasks.
$\mathcal{M}_{first,i}$ is the model's performance on the first task after learning $i$-th task, which is equivalent to $\mathcal{M}_{1,i}$.
$\mathcal{M}_{first}$ denotes the average performance of the model on the first task at the end of leaning each task.
$\mathcal{M}$ can be any of the evaluation metrics mentioned above.
$\mathcal{M}_{seen}$ evaluates model's overall performance on historical tasks, while $\mathcal{M}_{first}$ reflects the ability to avoid catastrophic forgetting.

\subsection{Baseline Methods}\label{subsec:baseline}

As we mentioned in Section~\ref{sec_intro} and Section~\ref{subsec_ll_of_qg}, none of the existing QG models can create questions across datasets and formats, since they are dedicated to specific tasks.
For the purposes of comparison, we train a Multitask-QG, a Finetuned-QG, and a Random-selected QG as our baseline.
All of the baselines utilize the same model architecture and QG encoding as our Unified-QG.

For Multitask-QG, the model is trained in a multi-task way. 
That is to say, the QG model is trained on a current dataset and \emph{all} the historical data, for each new task.
It can achieve the ``upper-bound'' performance of overall continual lifelong learning since it preserves and is trained on all current and historical data.

For Finetuned-QG, the model is trained in a transfer learning style.
With the progress of the tasks, the model is initialized with the checkpoint obtained from the last task, and then, it is finetuned with data from the current task.
This method might perform well on the current task but will suffer catastrophic forgetting.

Random-selected QG selects the examples from historical data uniformly. 
The advantage of the random-selected scheme is that it is simple and it can keep the selected samples and original data the same distribution to a large extent.

Note that all of the baseline methods utilize our proposed unified QG encoding method so that they can work across formats.
And all of the hyper-parameters settings for these baselines are the same as our Unified-QG. 
The detailed implementation is in Appendix~\ref{sec_imp_details}

\subsection{Overall Evaluation Results}\label{subsub_op}
The top half of Table~\ref{tb_all_performance} presents the results in terms of $M_{seen}$ which reflects the overall performance on all the datasets.
Finetuned-QG has the worst performance among all the comparison methods, since it suffers from the catastrophic forgetting problem.
Random-selected QG also has relatively poor performance, which shows the superiority of our difficulty-based sampling method.
As mentioned in Section~\ref{subsec:baseline}, Multitask-QG can be viewed as the ``upper-bound'' with regarding to $\mathcal{M}_{seen}$, since it stores all historical data and costs much more GPU computations to train the model.
Our Unified-QG achieves very close performance to the Multitask-QG, with much less computing and storage resources.

  \begin{table}[!ht]
    \renewcommand{\arraystretch}{0.9}
    \setlength\tabcolsep{2pt}
    \centering
    \caption{The comparison of each model's average continual learning performance ($\mathcal{M}_{seen}$ or $\mathcal{M}_{first}$). * means the ``upper-bound'' under the metrics $\mathcal{M}_{seen}$.}
    \begin{tabular}{l|cccccc}
    \hline
    \textbf{$\mathcal{M}_{seen}$}         & \textbf{B1} & \textbf{B2} & \textbf{B3} & \textbf{B4} & \textbf{M.} & \textbf{RL} \\ \hline
    \textbf{Finetuned-QG}                 & 32.58           & 20.04           & 13.84           & 10.02           & 17.83           & 34.39            \\
    \textbf{Random-selected QG}           & 40.93           & 26.71           & 19.18           & 14.34           & 20.66           & 41.20            \\
    \textbf{Multitask-QG*} & 42.31           & 28.31           & 20.70           & 15.67           & 21.56           & 42.85            \\ 
    \textbf{Unified-QG (ours)}            & \textbf{42.15}  & \textbf{28.17}  & \textbf{20.58}  & \textbf{15.69}  & \textbf{21.49}  & \textbf{42.79}   \\ \hline
    \textbf{$\mathcal{M}_{first}$}        & \textbf{B1} & \textbf{B2} & \textbf{B3} & \textbf{B4} & \textbf{M.} & \textbf{RL} \\ \hline
    \textbf{Finetuned-QG}                 & 34.97          &  21.37           & 14.56           & 10.27            & 20.02           & 37.62            \\
    \textbf{Random-selected QG}           & 48.63          & 33.78           & 25.83           & 20.40           & 24.26           & 48.92                 \\
    \textbf{Multitask-QG}                 & 47.84           & 33.72           & 25.89           & 20.36           & 24.32           & 48.76            \\ 
    \textbf{Unified-QG (ours)}            & \textbf{51.22}  & \textbf{36.61}  & \textbf{28.42}  & \textbf{22.54}  & \textbf{25.92}  & \textbf{51.79}   \\ \hline
    \end{tabular}
    \label{tb_all_performance}
    \end{table}

  \begin{figure}[!htbp]
    \centering
    \subfloat[The trend of average BLEU-4 scores on all seen datasets.]{\includegraphics[width=1.7in]{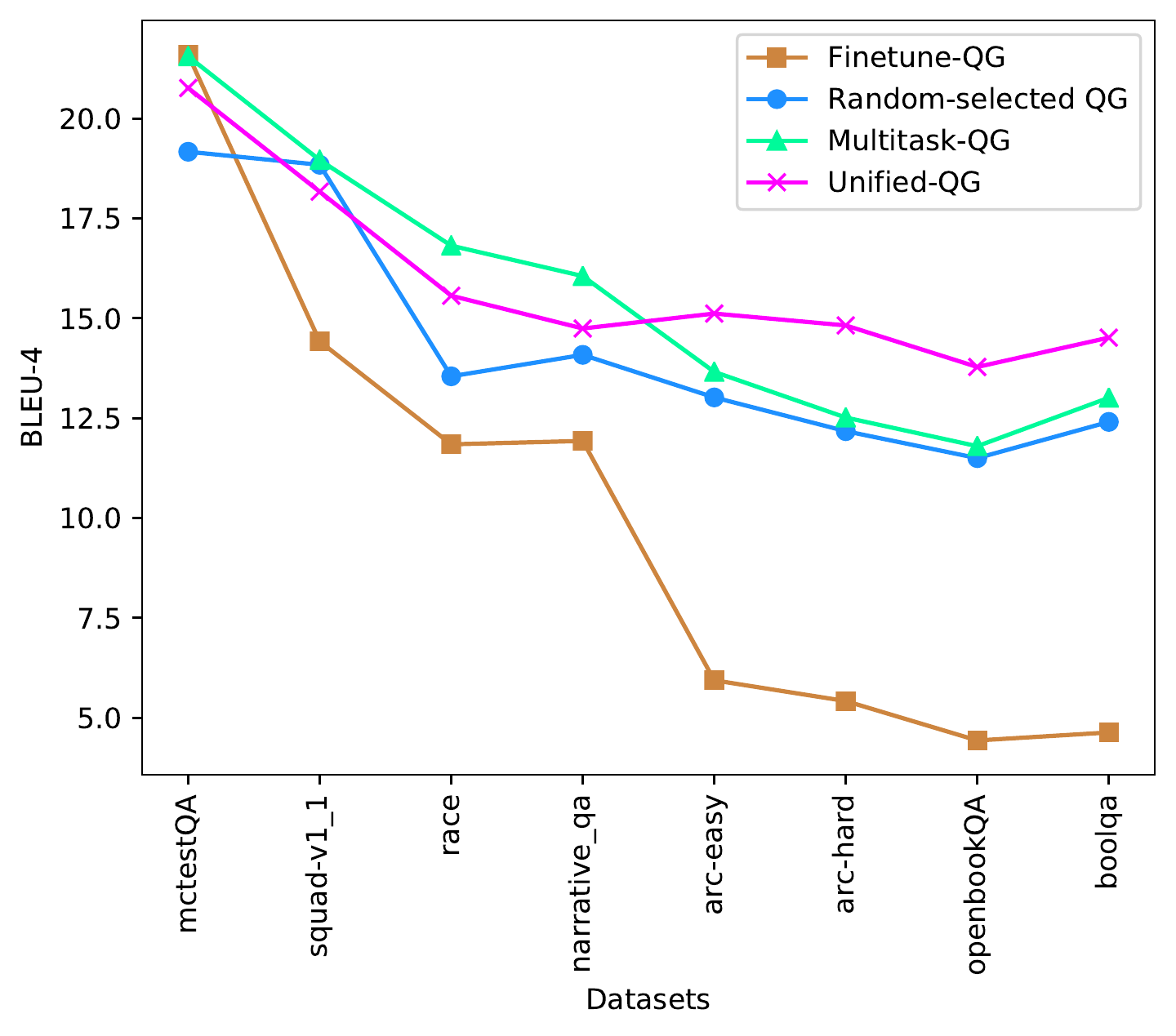}\label{fig_avg_bleu4_all_seen}}
    \subfloat[The trend of BLEU-4 scores on the first dataset.]{\includegraphics[width=1.65in]{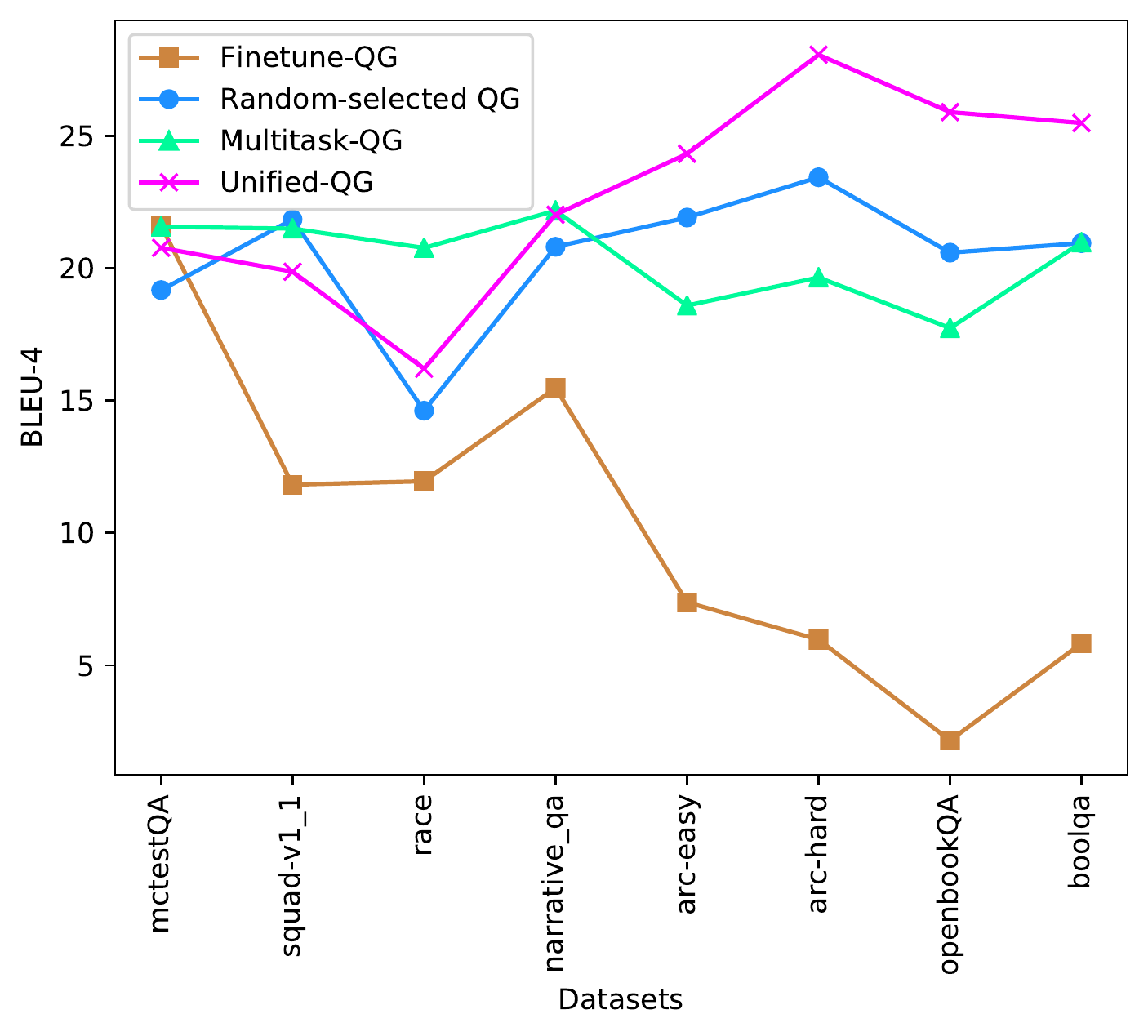}\label{fig_avg_bleu4_squad}}
    \caption{Detailed performance with $\mathcal{M}_{seen}$ and $\mathcal{M}_{first}$}
  \end{figure}
  
Figure~\ref{fig_avg_bleu4_all_seen} further shows the trend of BLEU-4 scores on all seen datasets (BLEU-$4_{seen,i}$).
We can observe that all the methods achieve comparable performance on the very front tasks, but the finetune method falls far behind the other methods from the third task due to its catastrophic forgetting nature.
The multitask method and random-selected method show similar performance, in fact, the latter one can be viewed as simplified version of the multitask method.
Our Unified-QG achieves better performance after learning $4$ tasks, exhibiting the strong ability to continually learn QG tasks.
The reason that Multitask-QG underperforms Unified-QG in the last four datasets is that it suffers from the significant discrepancy in data sizes.
Table~\ref{tb_detail_last_model} further shows that the final Multitask-QG only achieves good performance on the largest dataset RACE, and consistently underperforms on all small datasets.

\subsection{Effectiveness of Overcoming Catastrophic Forgetting Problem}\label{subsub_ef}
As we mentioned in Section~\ref{subsec_eval_metrics}, $\mathcal{M}_{first}$ measures the effectiveness of overcoming the forgetting problem. 
The bottom half of Table~\ref{tb_all_performance} shows each approach's average performance on the first dataset (McTestQA), which reflects the effectiveness of mitigating catastrophic forgetting.
As illustrated in Table~\ref{tb_all_performance}, simply finetune the QG model on each new dataset leads to severe forgetting, in other words, it indicates that catastrophic forgetting exists in continual learning QG.
Randomly select examples from the previous data slightly alleviates forgetting but still obtains relatively poor performance.
Multitask-QG performs best among all baselines, however, it utilizes all historical data when training on a new task, costing much more computation and storage resources and much longer training time.
Our Unified-QG outperforms Multitask-QG significantly and consistently in terms of both effectiveness and efficiency. 

Figure~\ref{fig_avg_bleu4_squad} shows the trend of BLEU-4 scores on the first dataset during the continual learning.
It reflects the amount of knowledge regarding the first task the model remembers as the learning continues.
From Figure~\ref{fig_avg_bleu4_squad}, we can observe that our Unified-QG's performance on McTestQA is improved by the later learned knowledge from other tasks.
  This interesting observation indicates that the underlying context understanding and question construction knowledge from latter tasks can be useful for previous tasks.
\begin{table}[!t]
  \renewcommand{\arraystretch}{0.9}
  \setlength\tabcolsep{4pt}
  \centering
  \caption{The detail performance of the last model on all the datasets. At each cell, the first line of value is for Finetuned-QG, the second line is for Random-selected QG, the third line is for Multitask-QG, and the last line is for our method. The best value is in bold.}
  \begin{tabular}{l|cccccc}
  \hline
  \textbf{Dataset}     & \textbf{B1}                                                   & \textbf{B2}                                                   & \textbf{B3}                                                   & \textbf{B4}                                                   & \textbf{M.}                                                   & \textbf{RL}                                                  \\ \hline
  \textbf{McTest}       & \begin{tabular}[c]{@{}c@{}}31.41\\ 50.21\\ 48.59\\ \textbf{54.35}\end{tabular} & \begin{tabular}[c]{@{}c@{}}15.76\\ 34.83\\ 34.92\\ \textbf{40.08}\end{tabular}  & \begin{tabular}[c]{@{}c@{}}9.40\\ 26.69\\ 26.77\\ \textbf{32.09}\end{tabular}  & \begin{tabular}[c]{@{}c@{}}5.82\\ 20.94\\ 20.96\\ \textbf{25.48}\end{tabular}  & \begin{tabular}[c]{@{}c@{}}17.43\\ 24.70\\ 24.87\\ \textbf{27.75}\end{tabular} & \begin{tabular}[c]{@{}c@{}}30.51\\ 50.35\\ 48.76\\ \textbf{55.54}\end{tabular} \\ \hline
  \textbf{SQuAD} & \begin{tabular}[c]{@{}c@{}}23.59\\ 44.66\\ 45.64\\ \textbf{46.50}\end{tabular} & \begin{tabular}[c]{@{}c@{}}8.68\\ 29.13\\ 30.10\\ \textbf{32.29}\end{tabular}  & \begin{tabular}[c]{@{}c@{}}4.02\\ 20.94\\ 21.75\\ \textbf{24.44}\end{tabular}  & \begin{tabular}[c]{@{}c@{}}2.058\\ 15.61\\ 16.30\\ \textbf{19.20}\end{tabular}  & \begin{tabular}[c]{@{}c@{}}11.56\\ 22.41\\ 23.06\\ \textbf{23.90}\end{tabular}  & \begin{tabular}[c]{@{}c@{}}23.02\\ 43.91\\ 44.56\\ \textbf{47.42}\end{tabular} \\ \hline
  \textbf{RACE}        & \begin{tabular}[c]{@{}c@{}}16.35\\ 36.51\\ \textbf{40.71}\\ 36.59\end{tabular} & \begin{tabular}[c]{@{}c@{}}6.24\\ 24.08\\ \textbf{28.28}\\ 24.86\end{tabular}  & \begin{tabular}[c]{@{}c@{}}3.00\\ 16.58\\ \textbf{20.36}\\ 17.71\end{tabular}  & \begin{tabular}[c]{@{}c@{}}1.63\\ 11.69\\ \textbf{15.00}\\ 12.95\end{tabular}  & \begin{tabular}[c]{@{}c@{}}8.58\\ 18.75\\ \textbf{21.34}\\ 19.33\end{tabular}  & \begin{tabular}[c]{@{}c@{}}14.23\\ 32.57\\ \textbf{37.09}\\ 33.53\end{tabular} \\ \hline
  \textbf{NarrativeQA}      & \begin{tabular}[c]{@{}c@{}}20.59\\ 37.84\\ 38.57\\ \textbf{38.64}\end{tabular} & \begin{tabular}[c]{@{}c@{}}5.16\\ 21.23\\ 22.13\\ \textbf{23.45}\end{tabular} & \begin{tabular}[c]{@{}c@{}}1.89\\ 14.02\\ 14.81\\ \textbf{16.43}\end{tabular}  & \begin{tabular}[c]{@{}c@{}}0.86\\ 9.83\\ 10.51\\ \textbf{12.19}\end{tabular}  & \begin{tabular}[c]{@{}c@{}}9.30\\ 17.51\\ 18.07\\ \textbf{18.64}\end{tabular} & \begin{tabular}[c]{@{}c@{}}20.30\\ 38.47\\ 39.37\\ \textbf{41.59}\end{tabular} \\ \hline
  \textbf{Arc-easy}    & \begin{tabular}[c]{@{}c@{}}16.56\\ \textbf{31.59}\\ 29.01\\ 30.85\end{tabular} & \begin{tabular}[c]{@{}c@{}}8.54\\ 19.61\\ 17.81\\ \textbf{19.81}\end{tabular}  & \begin{tabular}[c]{@{}c@{}}5.03\\ 13.34\\ 11.98\\ \textbf{13.85}\end{tabular}  & \begin{tabular}[c]{@{}c@{}}3.13\\ 9.50\\ 8.43\\ \textbf{10.17}\end{tabular}    & \begin{tabular}[c]{@{}c@{}}10.62\\ 17.32\\ 16.33\\ \textbf{17.78}\end{tabular} & \begin{tabular}[c]{@{}c@{}}24.36\\ 35.35\\ 34.14\\ \textbf{36.79}\end{tabular} \\ \hline
  \textbf{Arc-hard}    & \begin{tabular}[c]{@{}c@{}}12.05\\ 28.18\\ 25.33\\ \textbf{27.04}\end{tabular} & \begin{tabular}[c]{@{}c@{}}5.74\\ 16.28\\ 14.63\\ \textbf{16.51}\end{tabular}  & \begin{tabular}[c]{@{}c@{}}3.09\\ 10.21\\ 9.34\\ \textbf{11.05}\end{tabular}   & \begin{tabular}[c]{@{}c@{}}1.78\\ 6.87\\ 6.39\\ \textbf{7.83}\end{tabular}    & \begin{tabular}[c]{@{}c@{}}9.10\\ 15.75\\ 14.57\\ \textbf{15.82}\end{tabular}  & \begin{tabular}[c]{@{}c@{}}21.94\\ 31.82\\ 31.47\\ \textbf{33.66}\end{tabular} \\ \hline
  \textbf{OpenbookQA}  & \begin{tabular}[c]{@{}c@{}}22.13\\ 30.83\\ \textbf{35.08}\\ 34.32\end{tabular} & \begin{tabular}[c]{@{}c@{}}10.93\\ 18.21\\ 21.68\\ \textbf{22.57}\end{tabular} & \begin{tabular}[c]{@{}c@{}}5.79\\ 11.47\\ 14.21\\ \textbf{15.76}\end{tabular}  & \begin{tabular}[c]{@{}c@{}}3.22\\ 7.62\\ 9.73\\ \textbf{11.33}\end{tabular}    & \begin{tabular}[c]{@{}c@{}}11.80\\ 18.81\\ 19.38\\ \textbf{19.43}\end{tabular} & \begin{tabular}[c]{@{}c@{}}20.67\\ 30.50\\ 33.34\\ \textbf{35.18}\end{tabular} \\ \hline
  \textbf{BoolQA}      & \begin{tabular}[c]{@{}c@{}}\textbf{49.49}\\ 47.79\\ 47.90\\ 48.84\end{tabular} & \begin{tabular}[c]{@{}c@{}}\textbf{33.70}\\ 31.86\\ 31.62\\ 33.10\end{tabular} & \begin{tabular}[c]{@{}c@{}}\textbf{24.71}\\ 23.06\\ 22.74\\ 24.19\end{tabular} & \begin{tabular}[c]{@{}c@{}}\textbf{18.51}\\ 17.14\\ 16.73\\ 18.04\end{tabular} & \begin{tabular}[c]{@{}c@{}}\textbf{23.68}\\ 22.58\\ 22.25\\ 23.27\end{tabular} & \begin{tabular}[c]{@{}c@{}}\textbf{47.04}\\ 44.96\\ 45.16\\ 46.08\end{tabular} \\ \hline
  \end{tabular}
  \label{tb_detail_last_model}
  \end{table}

  \begin{table}[]
    \renewcommand{\arraystretch}{0.9}
    \centering
    \caption{The comparison of Unified-QG and $8$ dedicated trained QG models.}
      \begin{tabular}{l|cc}
      \hline
      \textbf{\diagbox {Dataset}{B4}}     & \multicolumn{1}{l}{\textbf{Dedicated-QG}} & \multicolumn{1}{l}{\textbf{Unified-QG}} \\ \hline
      \textbf{McTest}      & 24.43                                     & \textbf{25.48}                          \\
      \textbf{SQuAD}       & 17.59                                     & \textbf{19.20}                          \\
      \textbf{RACE}        & \textbf{17.47}                            & 12.95                                   \\
      \textbf{NarrativeQA} & 11.05                                     & \textbf{12.19}                          \\
      \textbf{Arc-easy}    & 9.44                                      & \textbf{10.17}                          \\
      \textbf{Arc-hard}    & 7.49                                      & \textbf{7.83}                           \\
      \textbf{OpenbookQA}  & \textbf{13.86}                            & 11.33                                   \\
      \textbf{BoolQA}      & \textbf{22.32}                            & 18.04                                   \\ \hline
      \end{tabular}
      \label{tb_dedicated}
      \end{table}

Table~\ref{tb_detail_last_model} describes the last model's performance on each dataset. 
In each cell, the four values represent the performance of Finetuned-QG, Random-selected QG, Multitask-QG, and Unified-QG respectively.
From Table~\ref{tb_detail_last_model} we can find that:
(1) If we simply finetune the QG model task by task, at the end of training, the last model nearly forgets all the learned knowledge from previous tasks and only performs well on the last dataset.
That means, in order to achieve good performance on all previous datasets using the traditional finetuning method, we have to retain all the historical model checkpoints, which will cost enormous storage.
(2) Compared to Finetuned-QG, Random-selected QG can alleviate the forgetting issue to a limited extent.
(3) Multitask-QG costs more computation time and storage than our approach, however, it only has better performance on RACE.
(4) Our Unified-QG outperforms all of the baselines on six datasets with more efficient computation and storage.

\subsection{Unified-QG vs. Dedicated-QG}
We further compare our Unified-QG with dedicated trained QG models, i.e. for each dataset, we train a QG model.
The architecture and all the training hyper-parameters of these dedicated QG models are the same as our Unified-QG.

Table~\ref{tb_dedicated} reports the BLEU-4 results of these dedicated QG models and our Unified-QG.
Note that the ``Dedicated-QG'' does not refer to a single model, it is a generic term for $8$ QG models, each of which is well-trained on only one specific dataset.
Our Unified-QG outperforms the specifically trained QG on McTest, SQuAD, NarrativeQA, Arc-easy, and Arc-hard.
Moreover, combined with Table~\ref{tb_dataset_statistics} we can observe that our Unified-QG outperforms dedicated QG on most small-scale datasets, which means low resource QG tasks are benefitted from such continually learning.

\subsection{Ablation Study}
In Table~\ref{tb_ablation}, to understand the effects of different components, we compare STRIDER with several simplified versions, including without similarity-based adaptation (-ST), without difficulty-based example selection (-D), i.e. random selection, and without example replay (-ER).
We can observe that 
(1) example replay is most important. Without it, the performance drops significantly from $15.69$ and $22.54$ to $11.24$ and $11.64$ in terms of $BLEU$-$4_{seen}$ and $BLEU$-$4_{first}$;
(2) difficult example replay is more effective than uniformly selected examples;
(3) similarity-based regularization is also beneficial since the performance of STRIDER degrades without it.
\begin{table}[!h]
  \renewcommand{\arraystretch}{0.9}
  \centering
  \caption{Ablation study for STRIDER.}
  \begin{tabular}{lcc}
  \hline
                   & \textbf{$B4_{seen}$} & \textbf{$B4_{first}$} \\ \hline
  \textbf{STRIDER} & \textbf{15.69}       & \textbf{22.54}                 \\
  \textbf{-ST}     & 15.53                & 21.92                 \\
  \textbf{-D}      & 14.34                & 20.40                 \\
  \textbf{-ER}     & 11.24                & 11.64                 \\ \hline
  \end{tabular}
  \label{tb_ablation}
  \end{table}

In addition to the above evaluations, we also show Unified-QG improves $8$ QA systems' performance in Appendix~\ref{sec_qa_qg}.

\section{Conclusion}
In this paper, we propose Unified-QG, which can continually learn QG tasks across datasets and formats based on lifelong learning.
Specifically, our Unified-QG contains a T5-based unified QG model and a lifelong learning strategy STRIDER.
The T5-based unified QG model consists of a unified QG converting mechanism that converts multiple QG formats into text-in-text-out form, and a T5 model that processes the unified input.
The STRIDER includes a difficulty-based example replay and a similarity-based adaptive regularization to enable the model to continually learn how to produce questions.
To the best of our knowledge, it is the first time to construct a QG model simultaneously address different QG problems crossing format boundaries based on lifelong learning approaches.
We conduct extensive experiments and analyses on $8$ QG datasets with $4$ QG formats to demonstrate the effectiveness of our Unified-QG.
Finally, we apply our single Unified-QG to improve $8$ QA systems without any architectural change, which is infeasible for traditional QG models.

\section{Acknowledgments}
The work is partly supported by the National Key Research and Development Program of China (2021YFB1715600), Australian Research Council Future Fellowship (FT210100624) and Discovery Project (DP190101985).


\bibliographystyle{ACM-Reference-Format}
\bibliography{sample-base}


\begin{thebibliography}{89}


\ifx \showCODEN    \undefined \def \showCODEN     #1{\unskip}     \fi
\ifx \showDOI      \undefined \def \showDOI       #1{#1}\fi
\ifx \showISBNx    \undefined \def \showISBNx     #1{\unskip}     \fi
\ifx \showISBNxiii \undefined \def \showISBNxiii  #1{\unskip}     \fi
\ifx \showISSN     \undefined \def \showISSN      #1{\unskip}     \fi
\ifx \showLCCN     \undefined \def \showLCCN      #1{\unskip}     \fi
\ifx \shownote     \undefined \def \shownote      #1{#1}          \fi
\ifx \showarticletitle \undefined \def \showarticletitle #1{#1}   \fi
\ifx \showURL      \undefined \def \showURL       {\relax}        \fi
\providecommand\bibfield[2]{#2}
\providecommand\bibinfo[2]{#2}
\providecommand\natexlab[1]{#1}
\providecommand\showeprint[2][]{arXiv:#2}

\bibitem[\protect\citeauthoryear{Aljundi, Babiloni, Elhoseiny, Rohrbach, and
  Tuytelaars}{Aljundi et~al\mbox{.}}{2018}]%
        {aljundi2018memory}
\bibfield{author}{\bibinfo{person}{Rahaf Aljundi}, \bibinfo{person}{Francesca
  Babiloni}, \bibinfo{person}{Mohamed Elhoseiny}, \bibinfo{person}{Marcus
  Rohrbach}, {and} \bibinfo{person}{Tinne Tuytelaars}.}
  \bibinfo{year}{2018}\natexlab{}.
\newblock \showarticletitle{Memory aware synapses: Learning what (not) to
  forget}. In \bibinfo{booktitle}{\emph{Proceedings of the European Conference
  on Computer Vision (ECCV)}}. \bibinfo{pages}{139--154}.
\newblock


\bibitem[\protect\citeauthoryear{Ba, Kiros, and Hinton}{Ba
  et~al\mbox{.}}{2016}]%
        {ba2016layer}
\bibfield{author}{\bibinfo{person}{Jimmy~Lei Ba}, \bibinfo{person}{Jamie~Ryan
  Kiros}, {and} \bibinfo{person}{Geoffrey~E Hinton}.}
  \bibinfo{year}{2016}\natexlab{}.
\newblock \showarticletitle{Layer normalization}.
\newblock \bibinfo{journal}{\emph{arXiv preprint arXiv:1607.06450}}
  (\bibinfo{year}{2016}).
\newblock


\bibitem[\protect\citeauthoryear{Bahdanau, Cho, and Bengio}{Bahdanau
  et~al\mbox{.}}{2014}]%
        {bahdanau2014neural}
\bibfield{author}{\bibinfo{person}{Dzmitry Bahdanau},
  \bibinfo{person}{Kyunghyun Cho}, {and} \bibinfo{person}{Yoshua Bengio}.}
  \bibinfo{year}{2014}\natexlab{}.
\newblock \showarticletitle{Neural machine translation by jointly learning to
  align and translate}.
\newblock \bibinfo{journal}{\emph{arXiv preprint arXiv:1409.0473}}
  (\bibinfo{year}{2014}).
\newblock


\bibitem[\protect\citeauthoryear{Bajaj, Campos, Craswell, Deng, Gao, Liu,
  Majumder, McNamara, Mitra, Nguyen, et~al\mbox{.}}{Bajaj
  et~al\mbox{.}}{2016}]%
        {bajaj2016ms}
\bibfield{author}{\bibinfo{person}{Payal Bajaj}, \bibinfo{person}{Daniel
  Campos}, \bibinfo{person}{Nick Craswell}, \bibinfo{person}{Li Deng},
  \bibinfo{person}{Jianfeng Gao}, \bibinfo{person}{Xiaodong Liu},
  \bibinfo{person}{Rangan Majumder}, \bibinfo{person}{Andrew McNamara},
  \bibinfo{person}{Bhaskar Mitra}, \bibinfo{person}{Tri Nguyen},
  {et~al\mbox{.}}} \bibinfo{year}{2016}\natexlab{}.
\newblock \showarticletitle{Ms marco: A human generated machine reading
  comprehension dataset}.
\newblock \bibinfo{journal}{\emph{arXiv preprint arXiv:1611.09268}}
  (\bibinfo{year}{2016}).
\newblock


\bibitem[\protect\citeauthoryear{Bao, Dong, Wei, Wang, Yang, Liu, Wang, Gao,
  Piao, Zhou, et~al\mbox{.}}{Bao et~al\mbox{.}}{2020}]%
        {bao2020unilmv2}
\bibfield{author}{\bibinfo{person}{Hangbo Bao}, \bibinfo{person}{Li Dong},
  \bibinfo{person}{Furu Wei}, \bibinfo{person}{Wenhui Wang},
  \bibinfo{person}{Nan Yang}, \bibinfo{person}{Xiaodong Liu},
  \bibinfo{person}{Yu Wang}, \bibinfo{person}{Jianfeng Gao},
  \bibinfo{person}{Songhao Piao}, \bibinfo{person}{Ming Zhou}, {et~al\mbox{.}}}
  \bibinfo{year}{2020}\natexlab{}.
\newblock \showarticletitle{Unilmv2: Pseudo-masked language models for unified
  language model pre-training}. In \bibinfo{booktitle}{\emph{International
  Conference on Machine Learning}}. PMLR, \bibinfo{pages}{642--652}.
\newblock


\bibitem[\protect\citeauthoryear{Biesialska, Biesialska, and
  Costa-juss{\`a}}{Biesialska et~al\mbox{.}}{2020}]%
        {biesialska2020continual}
\bibfield{author}{\bibinfo{person}{Magdalena Biesialska},
  \bibinfo{person}{Katarzyna Biesialska}, {and} \bibinfo{person}{Marta~R
  Costa-juss{\`a}}.} \bibinfo{year}{2020}\natexlab{}.
\newblock \showarticletitle{Continual lifelong learning in natural language
  processing: A survey}.
\newblock \bibinfo{journal}{\emph{arXiv preprint arXiv:2012.09823}}
  (\bibinfo{year}{2020}).
\newblock


\bibitem[\protect\citeauthoryear{Callison-Burch, Osborne, and
  Koehn}{Callison-Burch et~al\mbox{.}}{2006}]%
        {callison2006re}
\bibfield{author}{\bibinfo{person}{Chris Callison-Burch},
  \bibinfo{person}{Miles Osborne}, {and} \bibinfo{person}{Philipp Koehn}.}
  \bibinfo{year}{2006}\natexlab{}.
\newblock \showarticletitle{Re-evaluating the role of BLEU in machine
  translation research}. In \bibinfo{booktitle}{\emph{11th Conference of the
  European Chapter of the Association for Computational Linguistics}}.
\newblock


\bibitem[\protect\citeauthoryear{Castro, Mar{\'\i}n-Jim{\'e}nez, Guil, Schmid,
  and Alahari}{Castro et~al\mbox{.}}{2018}]%
        {castro2018end}
\bibfield{author}{\bibinfo{person}{Francisco~M Castro},
  \bibinfo{person}{Manuel~J Mar{\'\i}n-Jim{\'e}nez},
  \bibinfo{person}{Nicol{\'a}s Guil}, \bibinfo{person}{Cordelia Schmid}, {and}
  \bibinfo{person}{Karteek Alahari}.} \bibinfo{year}{2018}\natexlab{}.
\newblock \showarticletitle{End-to-end incremental learning}. In
  \bibinfo{booktitle}{\emph{Proceedings of the European conference on computer
  vision (ECCV)}}. \bibinfo{pages}{233--248}.
\newblock


\bibitem[\protect\citeauthoryear{Ch and Saha}{Ch and Saha}{2018}]%
        {ch2018automatic}
\bibfield{author}{\bibinfo{person}{Dhawaleswar~Rao Ch} {and}
  \bibinfo{person}{Sujan~Kumar Saha}.} \bibinfo{year}{2018}\natexlab{}.
\newblock \showarticletitle{Automatic multiple choice question generation from
  text: A survey}.
\newblock \bibinfo{journal}{\emph{IEEE Transactions on Learning Technologies}}
  \bibinfo{volume}{13}, \bibinfo{number}{1} (\bibinfo{year}{2018}),
  \bibinfo{pages}{14--25}.
\newblock


\bibitem[\protect\citeauthoryear{Chali and Baghaee}{Chali and Baghaee}{2018}]%
        {chali2018automatic}
\bibfield{author}{\bibinfo{person}{Yllias Chali} {and} \bibinfo{person}{Tina
  Baghaee}.} \bibinfo{year}{2018}\natexlab{}.
\newblock \showarticletitle{Automatic opinion question generation}. In
  \bibinfo{booktitle}{\emph{Proceedings of the 11th International Conference on
  Natural Language Generation}}. \bibinfo{pages}{152--158}.
\newblock


\bibitem[\protect\citeauthoryear{Chan and Fan}{Chan and Fan}{2019a}]%
        {chan2019bert}
\bibfield{author}{\bibinfo{person}{Ying-Hong Chan} {and}
  \bibinfo{person}{Yao-Chung Fan}.} \bibinfo{year}{2019}\natexlab{a}.
\newblock \showarticletitle{BERT for question generation}. In
  \bibinfo{booktitle}{\emph{Proceedings of the 12th International Conference on
  Natural Language Generation}}. \bibinfo{pages}{173--177}.
\newblock


\bibitem[\protect\citeauthoryear{Chan and Fan}{Chan and Fan}{2019b}]%
        {chan2019recurrent}
\bibfield{author}{\bibinfo{person}{Ying-Hong Chan} {and}
  \bibinfo{person}{Yao-Chung Fan}.} \bibinfo{year}{2019}\natexlab{b}.
\newblock \showarticletitle{A recurrent BERT-based model for question
  generation}. In \bibinfo{booktitle}{\emph{Proceedings of the 2nd Workshop on
  Machine Reading for Question Answering}}. \bibinfo{pages}{154--162}.
\newblock


\bibitem[\protect\citeauthoryear{Chaudhry, Rohrbach, Elhoseiny, Ajanthan,
  Dokania, Torr, and Ranzato}{Chaudhry et~al\mbox{.}}{2019}]%
        {chaudhry2019continual}
\bibfield{author}{\bibinfo{person}{Arslan Chaudhry}, \bibinfo{person}{Marcus
  Rohrbach}, \bibinfo{person}{Mohamed Elhoseiny},
  \bibinfo{person}{Thalaiyasingam Ajanthan}, \bibinfo{person}{Puneet~K
  Dokania}, \bibinfo{person}{Philip~HS Torr}, {and} \bibinfo{person}{M
  Ranzato}.} \bibinfo{year}{2019}\natexlab{}.
\newblock \showarticletitle{Continual learning with tiny episodic memories}.
\newblock  (\bibinfo{year}{2019}).
\newblock


\bibitem[\protect\citeauthoryear{Clark, Lee, Chang, Kwiatkowski, Collins, and
  Toutanova}{Clark et~al\mbox{.}}{2019}]%
        {clark2019boolq}
\bibfield{author}{\bibinfo{person}{Christopher Clark}, \bibinfo{person}{Kenton
  Lee}, \bibinfo{person}{Ming-Wei Chang}, \bibinfo{person}{Tom Kwiatkowski},
  \bibinfo{person}{Michael Collins}, {and} \bibinfo{person}{Kristina
  Toutanova}.} \bibinfo{year}{2019}\natexlab{}.
\newblock \showarticletitle{BoolQ: Exploring the surprising difficulty of
  natural yes/no questions}.
\newblock \bibinfo{journal}{\emph{arXiv preprint arXiv:1905.10044}}
  (\bibinfo{year}{2019}).
\newblock


\bibitem[\protect\citeauthoryear{Clark, Cowhey, Etzioni, Khot, Sabharwal,
  Schoenick, and Tafjord}{Clark et~al\mbox{.}}{2018}]%
        {clark2018think}
\bibfield{author}{\bibinfo{person}{Peter Clark}, \bibinfo{person}{Isaac
  Cowhey}, \bibinfo{person}{Oren Etzioni}, \bibinfo{person}{Tushar Khot},
  \bibinfo{person}{Ashish Sabharwal}, \bibinfo{person}{Carissa Schoenick},
  {and} \bibinfo{person}{Oyvind Tafjord}.} \bibinfo{year}{2018}\natexlab{}.
\newblock \showarticletitle{Think you have solved question answering? try arc,
  the ai2 reasoning challenge}.
\newblock \bibinfo{journal}{\emph{arXiv preprint arXiv:1803.05457}}
  (\bibinfo{year}{2018}).
\newblock


\bibitem[\protect\citeauthoryear{Crawshaw}{Crawshaw}{2020}]%
        {crawshaw2020multi}
\bibfield{author}{\bibinfo{person}{Michael Crawshaw}.}
  \bibinfo{year}{2020}\natexlab{}.
\newblock \showarticletitle{Multi-task learning with deep neural networks: A
  survey}.
\newblock \bibinfo{journal}{\emph{arXiv preprint arXiv:2009.09796}}
  (\bibinfo{year}{2020}).
\newblock


\bibitem[\protect\citeauthoryear{Denkowski and Lavie}{Denkowski and
  Lavie}{2014}]%
        {denkowski2014meteor}
\bibfield{author}{\bibinfo{person}{Michael Denkowski} {and}
  \bibinfo{person}{Alon Lavie}.} \bibinfo{year}{2014}\natexlab{}.
\newblock \showarticletitle{Meteor universal: Language specific translation
  evaluation for any target language}. In \bibinfo{booktitle}{\emph{Proceedings
  of the ninth workshop on statistical machine translation}}.
  \bibinfo{pages}{376--380}.
\newblock


\bibitem[\protect\citeauthoryear{Dong, Yang, Wang, Wei, Liu, Wang, Gao, Zhou,
  and Hon}{Dong et~al\mbox{.}}{2019}]%
        {dong2019unified}
\bibfield{author}{\bibinfo{person}{Li Dong}, \bibinfo{person}{Nan Yang},
  \bibinfo{person}{Wenhui Wang}, \bibinfo{person}{Furu Wei},
  \bibinfo{person}{Xiaodong Liu}, \bibinfo{person}{Yu Wang},
  \bibinfo{person}{Jianfeng Gao}, \bibinfo{person}{Ming Zhou}, {and}
  \bibinfo{person}{Hsiao-Wuen Hon}.} \bibinfo{year}{2019}\natexlab{}.
\newblock \showarticletitle{Unified language model pre-training for natural
  language understanding and generation}.
\newblock \bibinfo{journal}{\emph{arXiv preprint arXiv:1905.03197}}
  (\bibinfo{year}{2019}).
\newblock


\bibitem[\protect\citeauthoryear{Du, Shao, and Cardie}{Du
  et~al\mbox{.}}{2017}]%
        {du2017learning}
\bibfield{author}{\bibinfo{person}{Xinya Du}, \bibinfo{person}{Junru Shao},
  {and} \bibinfo{person}{Claire Cardie}.} \bibinfo{year}{2017}\natexlab{}.
\newblock \showarticletitle{Learning to ask: Neural question generation for
  reading comprehension}.
\newblock \bibinfo{journal}{\emph{arXiv preprint arXiv:1705.00106}}
  (\bibinfo{year}{2017}).
\newblock


\bibitem[\protect\citeauthoryear{Duan, Tang, Chen, and Zhou}{Duan
  et~al\mbox{.}}{2017}]%
        {duan2017question}
\bibfield{author}{\bibinfo{person}{Nan Duan}, \bibinfo{person}{Duyu Tang},
  \bibinfo{person}{Peng Chen}, {and} \bibinfo{person}{Ming Zhou}.}
  \bibinfo{year}{2017}\natexlab{}.
\newblock \showarticletitle{Question generation for question answering}. In
  \bibinfo{booktitle}{\emph{Proceedings of the 2017 Conference on Empirical
  Methods in Natural Language Processing}}. \bibinfo{pages}{866--874}.
\newblock


\bibitem[\protect\citeauthoryear{Fang, Wang, Gan, Sun, Liu, and Zhu}{Fang
  et~al\mbox{.}}{2020}]%
        {fang2020accelerating}
\bibfield{author}{\bibinfo{person}{Yuwei Fang}, \bibinfo{person}{Shuohang
  Wang}, \bibinfo{person}{Zhe Gan}, \bibinfo{person}{Siqi Sun},
  \bibinfo{person}{Jingjing Liu}, {and} \bibinfo{person}{Chenguang Zhu}.}
  \bibinfo{year}{2020}\natexlab{}.
\newblock \showarticletitle{Accelerating real-time question answering via
  question generation}.
\newblock \bibinfo{journal}{\emph{arXiv preprint arXiv:2009.05167}}
  (\bibinfo{year}{2020}).
\newblock


\bibitem[\protect\citeauthoryear{French}{French}{1999}]%
        {french1999catastrophic}
\bibfield{author}{\bibinfo{person}{Robert~M French}.}
  \bibinfo{year}{1999}\natexlab{}.
\newblock \showarticletitle{Catastrophic forgetting in connectionist networks}.
\newblock \bibinfo{journal}{\emph{Trends in cognitive sciences}}
  \bibinfo{volume}{3}, \bibinfo{number}{4} (\bibinfo{year}{1999}),
  \bibinfo{pages}{128--135}.
\newblock


\bibitem[\protect\citeauthoryear{Gulcehre, Ahn, Nallapati, Zhou, and
  Bengio}{Gulcehre et~al\mbox{.}}{2016}]%
        {gulcehre2016pointing}
\bibfield{author}{\bibinfo{person}{Caglar Gulcehre}, \bibinfo{person}{Sungjin
  Ahn}, \bibinfo{person}{Ramesh Nallapati}, \bibinfo{person}{Bowen Zhou}, {and}
  \bibinfo{person}{Yoshua Bengio}.} \bibinfo{year}{2016}\natexlab{}.
\newblock \showarticletitle{Pointing the unknown words}.
\newblock \bibinfo{journal}{\emph{arXiv preprint arXiv:1603.08148}}
  (\bibinfo{year}{2016}).
\newblock


\bibitem[\protect\citeauthoryear{Guo, Yin, Wang, Chen, Zhou, and Quoc
  Viet~Hung}{Guo et~al\mbox{.}}{2019}]%
        {guo2019streaming}
\bibfield{author}{\bibinfo{person}{Lei Guo}, \bibinfo{person}{Hongzhi Yin},
  \bibinfo{person}{Qinyong Wang}, \bibinfo{person}{Tong Chen},
  \bibinfo{person}{Alexander Zhou}, {and} \bibinfo{person}{Nguyen Quoc
  Viet~Hung}.} \bibinfo{year}{2019}\natexlab{}.
\newblock \showarticletitle{Streaming session-based recommendation}. In
  \bibinfo{booktitle}{\emph{Proceedings of the 25th ACM SIGKDD International
  Conference on Knowledge Discovery \& Data Mining}}.
  \bibinfo{pages}{1569--1577}.
\newblock


\bibitem[\protect\citeauthoryear{Han, Karunasekera, and Leckie}{Han
  et~al\mbox{.}}{2020}]%
        {han2020graph}
\bibfield{author}{\bibinfo{person}{Yi Han}, \bibinfo{person}{Shanika
  Karunasekera}, {and} \bibinfo{person}{Christopher Leckie}.}
  \bibinfo{year}{2020}\natexlab{}.
\newblock \showarticletitle{Graph neural networks with continual learning for
  fake news detection from social media}.
\newblock \bibinfo{journal}{\emph{arXiv preprint arXiv:2007.03316}}
  (\bibinfo{year}{2020}).
\newblock


\bibitem[\protect\citeauthoryear{He, Zhang, Ren, and Sun}{He
  et~al\mbox{.}}{2016}]%
        {he2016deep}
\bibfield{author}{\bibinfo{person}{Kaiming He}, \bibinfo{person}{Xiangyu
  Zhang}, \bibinfo{person}{Shaoqing Ren}, {and} \bibinfo{person}{Jian Sun}.}
  \bibinfo{year}{2016}\natexlab{}.
\newblock \showarticletitle{Deep residual learning for image recognition}. In
  \bibinfo{booktitle}{\emph{Proceedings of the IEEE conference on computer
  vision and pattern recognition}}. \bibinfo{pages}{770--778}.
\newblock


\bibitem[\protect\citeauthoryear{He, Zhao, and Chu}{He et~al\mbox{.}}{2021b}]%
        {he2021automl}
\bibfield{author}{\bibinfo{person}{Xin He}, \bibinfo{person}{Kaiyong Zhao},
  {and} \bibinfo{person}{Xiaowen Chu}.} \bibinfo{year}{2021}\natexlab{b}.
\newblock \showarticletitle{AutoML: A Survey of the State-of-the-Art}.
\newblock \bibinfo{journal}{\emph{Knowledge-Based Systems}}
  \bibinfo{volume}{212} (\bibinfo{year}{2021}), \bibinfo{pages}{106622}.
\newblock


\bibitem[\protect\citeauthoryear{He, Chen, Wu, Yuan, and Wu}{He
  et~al\mbox{.}}{2021a}]%
        {he2021unsupervised}
\bibfield{author}{\bibinfo{person}{Yi He}, \bibinfo{person}{Sheng Chen},
  \bibinfo{person}{Baijun Wu}, \bibinfo{person}{Xu Yuan}, {and}
  \bibinfo{person}{Xindong Wu}.} \bibinfo{year}{2021}\natexlab{a}.
\newblock \showarticletitle{Unsupervised Lifelong Learning with Curricula}. In
  \bibinfo{booktitle}{\emph{Proceedings of the Web Conference 2021}}.
  \bibinfo{pages}{3534--3545}.
\newblock


\bibitem[\protect\citeauthoryear{Hou, Pan, Loy, Wang, and Lin}{Hou
  et~al\mbox{.}}{2019}]%
        {hou2019learning}
\bibfield{author}{\bibinfo{person}{Saihui Hou}, \bibinfo{person}{Xinyu Pan},
  \bibinfo{person}{Chen~Change Loy}, \bibinfo{person}{Zilei Wang}, {and}
  \bibinfo{person}{Dahua Lin}.} \bibinfo{year}{2019}\natexlab{}.
\newblock \showarticletitle{Learning a unified classifier incrementally via
  rebalancing}. In \bibinfo{booktitle}{\emph{Proceedings of the IEEE/CVF
  Conference on Computer Vision and Pattern Recognition}}.
  \bibinfo{pages}{831--839}.
\newblock


\bibitem[\protect\citeauthoryear{Hu, Liu, Ma, Zhao, and Yan}{Hu
  et~al\mbox{.}}{2018}]%
        {hu2018aspect}
\bibfield{author}{\bibinfo{person}{Wenpeng Hu}, \bibinfo{person}{Bing Liu},
  \bibinfo{person}{Jinwen Ma}, \bibinfo{person}{Dongyan Zhao}, {and}
  \bibinfo{person}{Rui Yan}.} \bibinfo{year}{2018}\natexlab{}.
\newblock \showarticletitle{Aspect-based question generation}.
\newblock  (\bibinfo{year}{2018}).
\newblock


\bibitem[\protect\citeauthoryear{Ke, Liu, and Huang}{Ke et~al\mbox{.}}{2020}]%
        {ke2020continual}
\bibfield{author}{\bibinfo{person}{Zixuan Ke}, \bibinfo{person}{Bing Liu},
  {and} \bibinfo{person}{Xingchang Huang}.} \bibinfo{year}{2020}\natexlab{}.
\newblock \showarticletitle{Continual learning of a mixed sequence of similar
  and dissimilar tasks}.
\newblock \bibinfo{journal}{\emph{Advances in Neural Information Processing
  Systems}}  \bibinfo{volume}{33} (\bibinfo{year}{2020}).
\newblock


\bibitem[\protect\citeauthoryear{Ke, Xu, and Liu}{Ke et~al\mbox{.}}{2021}]%
        {ke2021adapting}
\bibfield{author}{\bibinfo{person}{Zixuan Ke}, \bibinfo{person}{Hu Xu}, {and}
  \bibinfo{person}{Bing Liu}.} \bibinfo{year}{2021}\natexlab{}.
\newblock \showarticletitle{Adapting BERT for Continual Learning of a Sequence
  of Aspect Sentiment Classification Tasks}. In
  \bibinfo{booktitle}{\emph{Proceedings of the 2021 Conference of the North
  American Chapter of the Association for Computational Linguistics: Human
  Language Technologies}}. \bibinfo{pages}{4746--4755}.
\newblock


\bibitem[\protect\citeauthoryear{Kemker and Kanan}{Kemker and Kanan}{2017}]%
        {kemker2017fearnet}
\bibfield{author}{\bibinfo{person}{Ronald Kemker} {and}
  \bibinfo{person}{Christopher Kanan}.} \bibinfo{year}{2017}\natexlab{}.
\newblock \showarticletitle{Fearnet: Brain-inspired model for incremental
  learning}.
\newblock \bibinfo{journal}{\emph{arXiv preprint arXiv:1711.10563}}
  (\bibinfo{year}{2017}).
\newblock


\bibitem[\protect\citeauthoryear{Kirkpatrick, Pascanu, Rabinowitz, Veness,
  Desjardins, Rusu, Milan, Quan, Ramalho, Grabska-Barwinska,
  et~al\mbox{.}}{Kirkpatrick et~al\mbox{.}}{2017}]%
        {kirkpatrick2017overcoming}
\bibfield{author}{\bibinfo{person}{James Kirkpatrick}, \bibinfo{person}{Razvan
  Pascanu}, \bibinfo{person}{Neil Rabinowitz}, \bibinfo{person}{Joel Veness},
  \bibinfo{person}{Guillaume Desjardins}, \bibinfo{person}{Andrei~A Rusu},
  \bibinfo{person}{Kieran Milan}, \bibinfo{person}{John Quan},
  \bibinfo{person}{Tiago Ramalho}, \bibinfo{person}{Agnieszka
  Grabska-Barwinska}, {et~al\mbox{.}}} \bibinfo{year}{2017}\natexlab{}.
\newblock \showarticletitle{Overcoming catastrophic forgetting in neural
  networks}.
\newblock \bibinfo{journal}{\emph{Proceedings of the national academy of
  sciences}} \bibinfo{volume}{114}, \bibinfo{number}{13}
  (\bibinfo{year}{2017}), \bibinfo{pages}{3521--3526}.
\newblock


\bibitem[\protect\citeauthoryear{Ko{\v{c}}isk{\`y}, Schwarz, Blunsom, Dyer,
  Hermann, Melis, and Grefenstette}{Ko{\v{c}}isk{\`y} et~al\mbox{.}}{2018}]%
        {kovcisky2018narrativeqa}
\bibfield{author}{\bibinfo{person}{Tom{\'a}{\v{s}} Ko{\v{c}}isk{\`y}},
  \bibinfo{person}{Jonathan Schwarz}, \bibinfo{person}{Phil Blunsom},
  \bibinfo{person}{Chris Dyer}, \bibinfo{person}{Karl~Moritz Hermann},
  \bibinfo{person}{G{\'a}bor Melis}, {and} \bibinfo{person}{Edward
  Grefenstette}.} \bibinfo{year}{2018}\natexlab{}.
\newblock \showarticletitle{The narrativeqa reading comprehension challenge}.
\newblock \bibinfo{journal}{\emph{Transactions of the Association for
  Computational Linguistics}}  \bibinfo{volume}{6} (\bibinfo{year}{2018}),
  \bibinfo{pages}{317--328}.
\newblock


\bibitem[\protect\citeauthoryear{Krishna and Iyyer}{Krishna and Iyyer}{2019}]%
        {krishna2019generating}
\bibfield{author}{\bibinfo{person}{Kalpesh Krishna} {and}
  \bibinfo{person}{Mohit Iyyer}.} \bibinfo{year}{2019}\natexlab{}.
\newblock \showarticletitle{Generating question-answer hierarchies}.
\newblock \bibinfo{journal}{\emph{arXiv preprint arXiv:1906.02622}}
  (\bibinfo{year}{2019}).
\newblock


\bibitem[\protect\citeauthoryear{Kwiatkowski, Palomaki, Redfield, Collins,
  Parikh, Alberti, Epstein, Polosukhin, Devlin, Lee, et~al\mbox{.}}{Kwiatkowski
  et~al\mbox{.}}{2019}]%
        {kwiatkowski2019natural}
\bibfield{author}{\bibinfo{person}{Tom Kwiatkowski},
  \bibinfo{person}{Jennimaria Palomaki}, \bibinfo{person}{Olivia Redfield},
  \bibinfo{person}{Michael Collins}, \bibinfo{person}{Ankur Parikh},
  \bibinfo{person}{Chris Alberti}, \bibinfo{person}{Danielle Epstein},
  \bibinfo{person}{Illia Polosukhin}, \bibinfo{person}{Jacob Devlin},
  \bibinfo{person}{Kenton Lee}, {et~al\mbox{.}}}
  \bibinfo{year}{2019}\natexlab{}.
\newblock \showarticletitle{Natural questions: a benchmark for question
  answering research}.
\newblock \bibinfo{journal}{\emph{Transactions of the Association for
  Computational Linguistics}}  \bibinfo{volume}{7} (\bibinfo{year}{2019}),
  \bibinfo{pages}{453--466}.
\newblock


\bibitem[\protect\citeauthoryear{Lai, Xie, Liu, Yang, and Hovy}{Lai
  et~al\mbox{.}}{2017}]%
        {lai2017race}
\bibfield{author}{\bibinfo{person}{Guokun Lai}, \bibinfo{person}{Qizhe Xie},
  \bibinfo{person}{Hanxiao Liu}, \bibinfo{person}{Yiming Yang}, {and}
  \bibinfo{person}{Eduard Hovy}.} \bibinfo{year}{2017}\natexlab{}.
\newblock \showarticletitle{Race: Large-scale reading comprehension dataset
  from examinations}.
\newblock \bibinfo{journal}{\emph{arXiv preprint arXiv:1704.04683}}
  (\bibinfo{year}{2017}).
\newblock


\bibitem[\protect\citeauthoryear{Lelkes, Tran, and Yu}{Lelkes
  et~al\mbox{.}}{2021}]%
        {lelkes2021quiz}
\bibfield{author}{\bibinfo{person}{Adam~D Lelkes}, \bibinfo{person}{Vinh~Q
  Tran}, {and} \bibinfo{person}{Cong Yu}.} \bibinfo{year}{2021}\natexlab{}.
\newblock \showarticletitle{Quiz-Style Question Generation for News Stories}.
  In \bibinfo{booktitle}{\emph{Proceedings of the Web Conference 2021}}.
  \bibinfo{pages}{2501--2511}.
\newblock


\bibitem[\protect\citeauthoryear{Li and Hoiem}{Li and Hoiem}{2017}]%
        {li2017learning}
\bibfield{author}{\bibinfo{person}{Zhizhong Li} {and} \bibinfo{person}{Derek
  Hoiem}.} \bibinfo{year}{2017}\natexlab{}.
\newblock \showarticletitle{Learning without forgetting}.
\newblock \bibinfo{journal}{\emph{IEEE transactions on pattern analysis and
  machine intelligence}} \bibinfo{volume}{40}, \bibinfo{number}{12}
  (\bibinfo{year}{2017}), \bibinfo{pages}{2935--2947}.
\newblock


\bibitem[\protect\citeauthoryear{Lin}{Lin}{2004}]%
        {lin2004rouge}
\bibfield{author}{\bibinfo{person}{Chin-Yew Lin}.}
  \bibinfo{year}{2004}\natexlab{}.
\newblock \showarticletitle{Rouge: A package for automatic evaluation of
  summaries}. In \bibinfo{booktitle}{\emph{Text summarization branches out}}.
  \bibinfo{pages}{74--81}.
\newblock


\bibitem[\protect\citeauthoryear{Lin and Och}{Lin and Och}{2004}]%
        {lin2004looking}
\bibfield{author}{\bibinfo{person}{Chin-Yew Lin} {and} \bibinfo{person}{FJ
  Och}.} \bibinfo{year}{2004}\natexlab{}.
\newblock \showarticletitle{Looking for a few good metrics: ROUGE and its
  evaluation}. In \bibinfo{booktitle}{\emph{Ntcir Workshop}}.
\newblock


\bibitem[\protect\citeauthoryear{Liu, Wei, Niu, Chen, and He}{Liu
  et~al\mbox{.}}{2020}]%
        {liu2020asking}
\bibfield{author}{\bibinfo{person}{Bang Liu}, \bibinfo{person}{Haojie Wei},
  \bibinfo{person}{Di Niu}, \bibinfo{person}{Haolan Chen}, {and}
  \bibinfo{person}{Yancheng He}.} \bibinfo{year}{2020}\natexlab{}.
\newblock \showarticletitle{Asking questions the human way: Scalable
  question-answer generation from text corpus}. In
  \bibinfo{booktitle}{\emph{Proceedings of The Web Conference 2020}}.
  \bibinfo{pages}{2032--2043}.
\newblock


\bibitem[\protect\citeauthoryear{Loshchilov and Hutter}{Loshchilov and
  Hutter}{2017}]%
        {loshchilov2017decoupled}
\bibfield{author}{\bibinfo{person}{Ilya Loshchilov} {and}
  \bibinfo{person}{Frank Hutter}.} \bibinfo{year}{2017}\natexlab{}.
\newblock \showarticletitle{Decoupled weight decay regularization}.
\newblock \bibinfo{journal}{\emph{arXiv preprint arXiv:1711.05101}}
  (\bibinfo{year}{2017}).
\newblock


\bibitem[\protect\citeauthoryear{Ma, Zhu, Zhou, and Li}{Ma
  et~al\mbox{.}}{2020}]%
        {ma2020improving}
\bibfield{author}{\bibinfo{person}{Xiyao Ma}, \bibinfo{person}{Qile Zhu},
  \bibinfo{person}{Yanlin Zhou}, {and} \bibinfo{person}{Xiaolin Li}.}
  \bibinfo{year}{2020}\natexlab{}.
\newblock \showarticletitle{Improving question generation with sentence-level
  semantic matching and answer position inferring}. In
  \bibinfo{booktitle}{\emph{Proceedings of the AAAI Conference on Artificial
  Intelligence}}, Vol.~\bibinfo{volume}{34}. \bibinfo{pages}{8464--8471}.
\newblock


\bibitem[\protect\citeauthoryear{Mancini, Ricci, Caputo, and Rota~Bulo}{Mancini
  et~al\mbox{.}}{2018}]%
        {mancini2018adding}
\bibfield{author}{\bibinfo{person}{Massimiliano Mancini},
  \bibinfo{person}{Elisa Ricci}, \bibinfo{person}{Barbara Caputo}, {and}
  \bibinfo{person}{Samuel Rota~Bulo}.} \bibinfo{year}{2018}\natexlab{}.
\newblock \showarticletitle{Adding new tasks to a single network with weight
  transformations using binary masks}. In \bibinfo{booktitle}{\emph{Proceedings
  of the European Conference on Computer Vision (ECCV) Workshops}}.
  \bibinfo{pages}{0--0}.
\newblock


\bibitem[\protect\citeauthoryear{McCloskey and Cohen}{McCloskey and
  Cohen}{1989}]%
        {mccloskey1989catastrophic}
\bibfield{author}{\bibinfo{person}{Michael McCloskey} {and}
  \bibinfo{person}{Neal~J Cohen}.} \bibinfo{year}{1989}\natexlab{}.
\newblock \showarticletitle{Catastrophic interference in connectionist
  networks: The sequential learning problem}.
\newblock In \bibinfo{booktitle}{\emph{Psychology of learning and motivation}}.
  Vol.~\bibinfo{volume}{24}. \bibinfo{publisher}{Elsevier},
  \bibinfo{pages}{109--165}.
\newblock


\bibitem[\protect\citeauthoryear{Mi, Chen, Zhao, Huang, and Faltings}{Mi
  et~al\mbox{.}}{2020}]%
        {mi2020continual}
\bibfield{author}{\bibinfo{person}{Fei Mi}, \bibinfo{person}{Liangwei Chen},
  \bibinfo{person}{Mengjie Zhao}, \bibinfo{person}{Minlie Huang}, {and}
  \bibinfo{person}{Boi Faltings}.} \bibinfo{year}{2020}\natexlab{}.
\newblock \showarticletitle{Continual learning for natural language generation
  in task-oriented dialog systems}.
\newblock \bibinfo{journal}{\emph{arXiv preprint arXiv:2010.00910}}
  (\bibinfo{year}{2020}).
\newblock


\bibitem[\protect\citeauthoryear{Mihaylov, Clark, Khot, and Sabharwal}{Mihaylov
  et~al\mbox{.}}{2018}]%
        {mihaylov2018can}
\bibfield{author}{\bibinfo{person}{Todor Mihaylov}, \bibinfo{person}{Peter
  Clark}, \bibinfo{person}{Tushar Khot}, {and} \bibinfo{person}{Ashish
  Sabharwal}.} \bibinfo{year}{2018}\natexlab{}.
\newblock \showarticletitle{Can a suit of armor conduct electricity? a new
  dataset for open book question answering}.
\newblock \bibinfo{journal}{\emph{arXiv preprint arXiv:1809.02789}}
  (\bibinfo{year}{2018}).
\newblock


\bibitem[\protect\citeauthoryear{Pan, Lei, Chua, and Kan}{Pan
  et~al\mbox{.}}{2019}]%
        {pan2019recent}
\bibfield{author}{\bibinfo{person}{Liangming Pan}, \bibinfo{person}{Wenqiang
  Lei}, \bibinfo{person}{Tat-Seng Chua}, {and} \bibinfo{person}{Min-Yen Kan}.}
  \bibinfo{year}{2019}\natexlab{}.
\newblock \showarticletitle{Recent advances in neural question generation}.
\newblock \bibinfo{journal}{\emph{arXiv preprint arXiv:1905.08949}}
  (\bibinfo{year}{2019}).
\newblock


\bibitem[\protect\citeauthoryear{Papineni, Roukos, Ward, and Zhu}{Papineni
  et~al\mbox{.}}{2002}]%
        {papineni2002bleu}
\bibfield{author}{\bibinfo{person}{Kishore Papineni}, \bibinfo{person}{Salim
  Roukos}, \bibinfo{person}{Todd Ward}, {and} \bibinfo{person}{Wei-Jing Zhu}.}
  \bibinfo{year}{2002}\natexlab{}.
\newblock \showarticletitle{Bleu: a method for automatic evaluation of machine
  translation}. In \bibinfo{booktitle}{\emph{Proceedings of the 40th annual
  meeting of the Association for Computational Linguistics}}.
  \bibinfo{pages}{311--318}.
\newblock


\bibitem[\protect\citeauthoryear{Parikh, T{\"a}ckstr{\"o}m, Das, and
  Uszkoreit}{Parikh et~al\mbox{.}}{2016}]%
        {parikh2016decomposable}
\bibfield{author}{\bibinfo{person}{Ankur~P Parikh}, \bibinfo{person}{Oscar
  T{\"a}ckstr{\"o}m}, \bibinfo{person}{Dipanjan Das}, {and}
  \bibinfo{person}{Jakob Uszkoreit}.} \bibinfo{year}{2016}\natexlab{}.
\newblock \showarticletitle{A decomposable attention model for natural language
  inference}.
\newblock \bibinfo{journal}{\emph{arXiv preprint arXiv:1606.01933}}
  (\bibinfo{year}{2016}).
\newblock


\bibitem[\protect\citeauthoryear{Parisi, Kemker, Part, Kanan, and
  Wermter}{Parisi et~al\mbox{.}}{2019}]%
        {parisi2019continual}
\bibfield{author}{\bibinfo{person}{German~I Parisi}, \bibinfo{person}{Ronald
  Kemker}, \bibinfo{person}{Jose~L Part}, \bibinfo{person}{Christopher Kanan},
  {and} \bibinfo{person}{Stefan Wermter}.} \bibinfo{year}{2019}\natexlab{}.
\newblock \showarticletitle{Continual lifelong learning with neural networks: A
  review}.
\newblock \bibinfo{journal}{\emph{Neural Networks}}  \bibinfo{volume}{113}
  (\bibinfo{year}{2019}), \bibinfo{pages}{54--71}.
\newblock


\bibitem[\protect\citeauthoryear{Peng, Zhu, Li, Li, Li, Zeng, and Gao}{Peng
  et~al\mbox{.}}{2020}]%
        {peng2020few}
\bibfield{author}{\bibinfo{person}{Baolin Peng}, \bibinfo{person}{Chenguang
  Zhu}, \bibinfo{person}{Chunyuan Li}, \bibinfo{person}{Xiujun Li},
  \bibinfo{person}{Jinchao Li}, \bibinfo{person}{Michael Zeng}, {and}
  \bibinfo{person}{Jianfeng Gao}.} \bibinfo{year}{2020}\natexlab{}.
\newblock \showarticletitle{Few-shot natural language generation for
  task-oriented dialog}.
\newblock \bibinfo{journal}{\emph{arXiv preprint arXiv:2002.12328}}
  (\bibinfo{year}{2020}).
\newblock


\bibitem[\protect\citeauthoryear{Qaiser and Ali}{Qaiser and Ali}{2018}]%
        {qaiser2018text}
\bibfield{author}{\bibinfo{person}{Shahzad Qaiser} {and}
  \bibinfo{person}{Ramsha Ali}.} \bibinfo{year}{2018}\natexlab{}.
\newblock \showarticletitle{Text mining: use of TF-IDF to examine the relevance
  of words to documents}.
\newblock \bibinfo{journal}{\emph{International Journal of Computer
  Applications}} \bibinfo{volume}{181}, \bibinfo{number}{1}
  (\bibinfo{year}{2018}), \bibinfo{pages}{25--29}.
\newblock


\bibitem[\protect\citeauthoryear{Qi, Zhang, and Manning}{Qi
  et~al\mbox{.}}{2020}]%
        {qi2020stay}
\bibfield{author}{\bibinfo{person}{Peng Qi}, \bibinfo{person}{Yuhao Zhang},
  {and} \bibinfo{person}{Christopher~D Manning}.}
  \bibinfo{year}{2020}\natexlab{}.
\newblock \showarticletitle{Stay hungry, stay focused: Generating informative
  and specific questions in information-seeking conversations}.
\newblock \bibinfo{journal}{\emph{arXiv preprint arXiv:2004.14530}}
  (\bibinfo{year}{2020}).
\newblock


\bibitem[\protect\citeauthoryear{Raffel, Shazeer, Roberts, Lee, Narang, Matena,
  Zhou, Li, and Liu}{Raffel et~al\mbox{.}}{2019}]%
        {raffel2019exploring}
\bibfield{author}{\bibinfo{person}{Colin Raffel}, \bibinfo{person}{Noam
  Shazeer}, \bibinfo{person}{Adam Roberts}, \bibinfo{person}{Katherine Lee},
  \bibinfo{person}{Sharan Narang}, \bibinfo{person}{Michael Matena},
  \bibinfo{person}{Yanqi Zhou}, \bibinfo{person}{Wei Li}, {and}
  \bibinfo{person}{Peter~J Liu}.} \bibinfo{year}{2019}\natexlab{}.
\newblock \showarticletitle{Exploring the limits of transfer learning with a
  unified text-to-text transformer}.
\newblock \bibinfo{journal}{\emph{arXiv preprint arXiv:1910.10683}}
  (\bibinfo{year}{2019}).
\newblock


\bibitem[\protect\citeauthoryear{Rajpurkar, Zhang, Lopyrev, and
  Liang}{Rajpurkar et~al\mbox{.}}{2016}]%
        {rajpurkar2016squad}
\bibfield{author}{\bibinfo{person}{Pranav Rajpurkar}, \bibinfo{person}{Jian
  Zhang}, \bibinfo{person}{Konstantin Lopyrev}, {and} \bibinfo{person}{Percy
  Liang}.} \bibinfo{year}{2016}\natexlab{}.
\newblock \showarticletitle{Squad: 100,000+ questions for machine comprehension
  of text}.
\newblock \bibinfo{journal}{\emph{arXiv preprint arXiv:1606.05250}}
  (\bibinfo{year}{2016}).
\newblock


\bibitem[\protect\citeauthoryear{Ramalho and Garnelo}{Ramalho and
  Garnelo}{2019}]%
        {ramalho2019adaptive}
\bibfield{author}{\bibinfo{person}{Tiago Ramalho} {and} \bibinfo{person}{Marta
  Garnelo}.} \bibinfo{year}{2019}\natexlab{}.
\newblock \showarticletitle{Adaptive posterior learning: few-shot learning with
  a surprise-based memory module}.
\newblock \bibinfo{journal}{\emph{arXiv preprint arXiv:1902.02527}}
  (\bibinfo{year}{2019}).
\newblock


\bibitem[\protect\citeauthoryear{Ramos et~al\mbox{.}}{Ramos
  et~al\mbox{.}}{2003}]%
        {ramos2003using}
\bibfield{author}{\bibinfo{person}{Juan Ramos} {et~al\mbox{.}}}
  \bibinfo{year}{2003}\natexlab{}.
\newblock \showarticletitle{Using tf-idf to determine word relevance in
  document queries}. In \bibinfo{booktitle}{\emph{Proceedings of the first
  instructional conference on machine learning}}, Vol.~\bibinfo{volume}{242}.
  Citeseer, \bibinfo{pages}{29--48}.
\newblock


\bibitem[\protect\citeauthoryear{Rebuffi, Kolesnikov, Sperl, and
  Lampert}{Rebuffi et~al\mbox{.}}{2017}]%
        {rebuffi2017icarl}
\bibfield{author}{\bibinfo{person}{Sylvestre-Alvise Rebuffi},
  \bibinfo{person}{Alexander Kolesnikov}, \bibinfo{person}{Georg Sperl}, {and}
  \bibinfo{person}{Christoph~H Lampert}.} \bibinfo{year}{2017}\natexlab{}.
\newblock \showarticletitle{icarl: Incremental classifier and representation
  learning}. In \bibinfo{booktitle}{\emph{Proceedings of the IEEE conference on
  Computer Vision and Pattern Recognition}}. \bibinfo{pages}{2001--2010}.
\newblock


\bibitem[\protect\citeauthoryear{Ren and Zhu}{Ren and Zhu}{2020}]%
        {ren2020knowledge}
\bibfield{author}{\bibinfo{person}{Siyu Ren} {and} \bibinfo{person}{Kenny~Q
  Zhu}.} \bibinfo{year}{2020}\natexlab{}.
\newblock \showarticletitle{Knowledge-Driven Distractor Generation for
  Cloze-Style Multiple Choice Questions}.
\newblock \bibinfo{journal}{\emph{arXiv preprint arXiv:2004.09853}}
  (\bibinfo{year}{2020}).
\newblock


\bibitem[\protect\citeauthoryear{Ren, Yin, Chen, Wang, Huang, and Zheng}{Ren
  et~al\mbox{.}}{2021}]%
        {ren2021learning}
\bibfield{author}{\bibinfo{person}{Xuhui Ren}, \bibinfo{person}{Hongzhi Yin},
  \bibinfo{person}{Tong Chen}, \bibinfo{person}{Hao Wang}, \bibinfo{person}{Zi
  Huang}, {and} \bibinfo{person}{Kai Zheng}.} \bibinfo{year}{2021}\natexlab{}.
\newblock \showarticletitle{Learning to Ask Appropriate Questions in
  Conversational Recommendation}.
\newblock \bibinfo{journal}{\emph{arXiv preprint arXiv:2105.04774}}
  (\bibinfo{year}{2021}).
\newblock


\bibitem[\protect\citeauthoryear{Richardson, Burges, and Renshaw}{Richardson
  et~al\mbox{.}}{2013}]%
        {richardson2013mctest}
\bibfield{author}{\bibinfo{person}{Matthew Richardson},
  \bibinfo{person}{Christopher~JC Burges}, {and} \bibinfo{person}{Erin
  Renshaw}.} \bibinfo{year}{2013}\natexlab{}.
\newblock \showarticletitle{Mctest: A challenge dataset for the open-domain
  machine comprehension of text}. In \bibinfo{booktitle}{\emph{Proceedings of
  the 2013 conference on empirical methods in natural language processing}}.
  \bibinfo{pages}{193--203}.
\newblock


\bibitem[\protect\citeauthoryear{Sennrich, Haddow, and Birch}{Sennrich
  et~al\mbox{.}}{2015}]%
        {sennrich2015improving}
\bibfield{author}{\bibinfo{person}{Rico Sennrich}, \bibinfo{person}{Barry
  Haddow}, {and} \bibinfo{person}{Alexandra Birch}.}
  \bibinfo{year}{2015}\natexlab{}.
\newblock \showarticletitle{Improving neural machine translation models with
  monolingual data}.
\newblock \bibinfo{journal}{\emph{arXiv preprint arXiv:1511.06709}}
  (\bibinfo{year}{2015}).
\newblock


\bibitem[\protect\citeauthoryear{Shin, Lee, Kim, and Kim}{Shin
  et~al\mbox{.}}{2017}]%
        {shin2017continual}
\bibfield{author}{\bibinfo{person}{Hanul Shin}, \bibinfo{person}{Jung~Kwon
  Lee}, \bibinfo{person}{Jaehong Kim}, {and} \bibinfo{person}{Jiwon Kim}.}
  \bibinfo{year}{2017}\natexlab{}.
\newblock \showarticletitle{Continual learning with deep generative replay}.
\newblock \bibinfo{journal}{\emph{arXiv preprint arXiv:1705.08690}}
  (\bibinfo{year}{2017}).
\newblock


\bibitem[\protect\citeauthoryear{Song, Cohn, and Specia}{Song
  et~al\mbox{.}}{2013}]%
        {song2013bleu}
\bibfield{author}{\bibinfo{person}{Xingyi Song}, \bibinfo{person}{Trevor Cohn},
  {and} \bibinfo{person}{Lucia Specia}.} \bibinfo{year}{2013}\natexlab{}.
\newblock \showarticletitle{BLEU deconstructed: Designing a better MT
  evaluation metric}.
\newblock \bibinfo{journal}{\emph{International Journal of Computational
  Linguistics and Applications}} \bibinfo{volume}{4}, \bibinfo{number}{2}
  (\bibinfo{year}{2013}), \bibinfo{pages}{29--44}.
\newblock


\bibitem[\protect\citeauthoryear{Srivastava, Hinton, Krizhevsky, Sutskever, and
  Salakhutdinov}{Srivastava et~al\mbox{.}}{2014}]%
        {srivastava2014dropout}
\bibfield{author}{\bibinfo{person}{Nitish Srivastava},
  \bibinfo{person}{Geoffrey Hinton}, \bibinfo{person}{Alex Krizhevsky},
  \bibinfo{person}{Ilya Sutskever}, {and} \bibinfo{person}{Ruslan
  Salakhutdinov}.} \bibinfo{year}{2014}\natexlab{}.
\newblock \showarticletitle{Dropout: a simple way to prevent neural networks
  from overfitting}.
\newblock \bibinfo{journal}{\emph{The journal of machine learning research}}
  \bibinfo{volume}{15}, \bibinfo{number}{1} (\bibinfo{year}{2014}),
  \bibinfo{pages}{1929--1958}.
\newblock


\bibitem[\protect\citeauthoryear{Sugiyama, Hatano, Yoshikawa, and
  Uemura}{Sugiyama et~al\mbox{.}}{2003}]%
        {sugiyama2003refinement}
\bibfield{author}{\bibinfo{person}{Kazunari Sugiyama}, \bibinfo{person}{Kenji
  Hatano}, \bibinfo{person}{Masatoshi Yoshikawa}, {and}
  \bibinfo{person}{Shunsuke Uemura}.} \bibinfo{year}{2003}\natexlab{}.
\newblock \showarticletitle{Refinement of TF-IDF schemes for web pages using
  their hyperlinked neighboring pages}. In
  \bibinfo{booktitle}{\emph{Proceedings of the fourteenth ACM conference on
  hypertext and hypermedia}}. \bibinfo{pages}{198--207}.
\newblock


\bibitem[\protect\citeauthoryear{Sun, Liu, Lyu, He, Ma, and Wang}{Sun
  et~al\mbox{.}}{2018}]%
        {sun2018answer}
\bibfield{author}{\bibinfo{person}{Xingwu Sun}, \bibinfo{person}{Jing Liu},
  \bibinfo{person}{Yajuan Lyu}, \bibinfo{person}{Wei He},
  \bibinfo{person}{Yanjun Ma}, {and} \bibinfo{person}{Shi Wang}.}
  \bibinfo{year}{2018}\natexlab{}.
\newblock \showarticletitle{Answer-focused and position-aware neural question
  generation}. In \bibinfo{booktitle}{\emph{Proceedings of the 2018 Conference
  on Empirical Methods in Natural Language Processing}}.
  \bibinfo{pages}{3930--3939}.
\newblock


\bibitem[\protect\citeauthoryear{Sun, Tang, Duan, Qin, Liu, Yan, Zhou, Lv, Yin,
  Feng, et~al\mbox{.}}{Sun et~al\mbox{.}}{2019}]%
        {sun2019joint}
\bibfield{author}{\bibinfo{person}{Yibo Sun}, \bibinfo{person}{Duyu Tang},
  \bibinfo{person}{Nan Duan}, \bibinfo{person}{Tao Qin},
  \bibinfo{person}{Shujie Liu}, \bibinfo{person}{Zhao Yan},
  \bibinfo{person}{Ming Zhou}, \bibinfo{person}{Yuanhua Lv},
  \bibinfo{person}{Wenpeng Yin}, \bibinfo{person}{Xiaocheng Feng},
  {et~al\mbox{.}}} \bibinfo{year}{2019}\natexlab{}.
\newblock \showarticletitle{Joint learning of question answering and question
  generation}.
\newblock \bibinfo{journal}{\emph{IEEE Transactions on Knowledge and Data
  Engineering}} \bibinfo{volume}{32}, \bibinfo{number}{5}
  (\bibinfo{year}{2019}), \bibinfo{pages}{971--982}.
\newblock


\bibitem[\protect\citeauthoryear{Thrun and Mitchell}{Thrun and
  Mitchell}{1995}]%
        {thrun1995lifelong}
\bibfield{author}{\bibinfo{person}{Sebastian Thrun} {and}
  \bibinfo{person}{Tom~M Mitchell}.} \bibinfo{year}{1995}\natexlab{}.
\newblock \showarticletitle{Lifelong robot learning}.
\newblock \bibinfo{journal}{\emph{Robotics and autonomous systems}}
  \bibinfo{volume}{15}, \bibinfo{number}{1-2} (\bibinfo{year}{1995}),
  \bibinfo{pages}{25--46}.
\newblock


\bibitem[\protect\citeauthoryear{Trischler, Wang, Yuan, Harris, Sordoni,
  Bachman, and Suleman}{Trischler et~al\mbox{.}}{2016}]%
        {trischler2016newsqa}
\bibfield{author}{\bibinfo{person}{Adam Trischler}, \bibinfo{person}{Tong
  Wang}, \bibinfo{person}{Xingdi Yuan}, \bibinfo{person}{Justin Harris},
  \bibinfo{person}{Alessandro Sordoni}, \bibinfo{person}{Philip Bachman}, {and}
  \bibinfo{person}{Kaheer Suleman}.} \bibinfo{year}{2016}\natexlab{}.
\newblock \showarticletitle{Newsqa: A machine comprehension dataset}.
\newblock \bibinfo{journal}{\emph{arXiv preprint arXiv:1611.09830}}
  (\bibinfo{year}{2016}).
\newblock


\bibitem[\protect\citeauthoryear{Vaswani, Shazeer, Parmar, Uszkoreit, Jones,
  Gomez, Kaiser, and Polosukhin}{Vaswani et~al\mbox{.}}{2017}]%
        {vaswani2017attention}
\bibfield{author}{\bibinfo{person}{Ashish Vaswani}, \bibinfo{person}{Noam
  Shazeer}, \bibinfo{person}{Niki Parmar}, \bibinfo{person}{Jakob Uszkoreit},
  \bibinfo{person}{Llion Jones}, \bibinfo{person}{Aidan~N Gomez},
  \bibinfo{person}{{\L}ukasz Kaiser}, {and} \bibinfo{person}{Illia
  Polosukhin}.} \bibinfo{year}{2017}\natexlab{}.
\newblock \showarticletitle{Attention is all you need}. In
  \bibinfo{booktitle}{\emph{Advances in neural information processing
  systems}}. \bibinfo{pages}{5998--6008}.
\newblock


\bibitem[\protect\citeauthoryear{Wang, Yin, Hu, Lian, Wang, and Huang}{Wang
  et~al\mbox{.}}{2018a}]%
        {wang2018neural}
\bibfield{author}{\bibinfo{person}{Qinyong Wang}, \bibinfo{person}{Hongzhi
  Yin}, \bibinfo{person}{Zhiting Hu}, \bibinfo{person}{Defu Lian},
  \bibinfo{person}{Hao Wang}, {and} \bibinfo{person}{Zi Huang}.}
  \bibinfo{year}{2018}\natexlab{a}.
\newblock \showarticletitle{Neural memory streaming recommender networks with
  adversarial training}. In \bibinfo{booktitle}{\emph{Proceedings of the 24th
  ACM SIGKDD International Conference on Knowledge Discovery \& Data Mining}}.
  \bibinfo{pages}{2467--2475}.
\newblock


\bibitem[\protect\citeauthoryear{Wang, Yin, Wang, Nguyen, Huang, and Cui}{Wang
  et~al\mbox{.}}{2019}]%
        {wang2019enhancing}
\bibfield{author}{\bibinfo{person}{Qinyong Wang}, \bibinfo{person}{Hongzhi
  Yin}, \bibinfo{person}{Hao Wang}, \bibinfo{person}{Quoc Viet~Hung Nguyen},
  \bibinfo{person}{Zi Huang}, {and} \bibinfo{person}{Lizhen Cui}.}
  \bibinfo{year}{2019}\natexlab{}.
\newblock \showarticletitle{Enhancing collaborative filtering with generative
  augmentation}. In \bibinfo{booktitle}{\emph{Proceedings of the 25th ACM
  SIGKDD International Conference on Knowledge Discovery \& Data Mining}}.
  \bibinfo{pages}{548--556}.
\newblock


\bibitem[\protect\citeauthoryear{Wang, Yin, Huang, Wang, Du, and Nguyen}{Wang
  et~al\mbox{.}}{2018b}]%
        {wang2018streaming}
\bibfield{author}{\bibinfo{person}{Weiqing Wang}, \bibinfo{person}{Hongzhi
  Yin}, \bibinfo{person}{Zi Huang}, \bibinfo{person}{Qinyong Wang},
  \bibinfo{person}{Xingzhong Du}, {and} \bibinfo{person}{Quoc Viet~Hung
  Nguyen}.} \bibinfo{year}{2018}\natexlab{b}.
\newblock \showarticletitle{Streaming ranking based recommender systems}. In
  \bibinfo{booktitle}{\emph{The 41st International ACM SIGIR Conference on
  Research \& Development in Information Retrieval}}.
  \bibinfo{pages}{525--534}.
\newblock


\bibitem[\protect\citeauthoryear{Welling}{Welling}{2009}]%
        {welling2009herding}
\bibfield{author}{\bibinfo{person}{Max Welling}.}
  \bibinfo{year}{2009}\natexlab{}.
\newblock \showarticletitle{Herding dynamical weights to learn}. In
  \bibinfo{booktitle}{\emph{Proceedings of the 26th Annual International
  Conference on Machine Learning}}. \bibinfo{pages}{1121--1128}.
\newblock


\bibitem[\protect\citeauthoryear{Yu, Quan, Su, and Yin}{Yu
  et~al\mbox{.}}{2020}]%
        {yu2020generating}
\bibfield{author}{\bibinfo{person}{Jianxing Yu}, \bibinfo{person}{Xiaojun
  Quan}, \bibinfo{person}{Qinliang Su}, {and} \bibinfo{person}{Jian Yin}.}
  \bibinfo{year}{2020}\natexlab{}.
\newblock \showarticletitle{Generating multi-hop reasoning questions to improve
  machine reading comprehension}. In \bibinfo{booktitle}{\emph{Proceedings of
  The Web Conference 2020}}. \bibinfo{pages}{281--291}.
\newblock


\bibitem[\protect\citeauthoryear{Yu, Bing, Zhang, Lam, and Si}{Yu
  et~al\mbox{.}}{2019}]%
        {yu2019based}
\bibfield{author}{\bibinfo{person}{Qian Yu}, \bibinfo{person}{Lidong Bing},
  \bibinfo{person}{Qiong Zhang}, \bibinfo{person}{Wai Lam}, {and}
  \bibinfo{person}{Luo Si}.} \bibinfo{year}{2019}\natexlab{}.
\newblock \showarticletitle{based Question Generation with Adaptive Instance
  Transfer and Augmentation}.
\newblock \bibinfo{journal}{\emph{arXiv preprint arXiv:1911.01556}}
  (\bibinfo{year}{2019}).
\newblock


\bibitem[\protect\citeauthoryear{Yuan, He, and Dai}{Yuan et~al\mbox{.}}{2021}]%
        {yuan2021improving}
\bibfield{author}{\bibinfo{person}{Wei Yuan}, \bibinfo{person}{Tieke He}, {and}
  \bibinfo{person}{Xinyu Dai}.} \bibinfo{year}{2021}\natexlab{}.
\newblock \showarticletitle{Improving Neural Question Generation using Deep
  Linguistic Representation}. In \bibinfo{booktitle}{\emph{Proceedings of the
  Web Conference 2021}}. \bibinfo{pages}{3489--3500}.
\newblock


\bibitem[\protect\citeauthoryear{Zhang, Zhao, Saleh, and Liu}{Zhang
  et~al\mbox{.}}{2020b}]%
        {zhang2020pegasus}
\bibfield{author}{\bibinfo{person}{Jingqing Zhang}, \bibinfo{person}{Yao Zhao},
  \bibinfo{person}{Mohammad Saleh}, {and} \bibinfo{person}{Peter Liu}.}
  \bibinfo{year}{2020}\natexlab{b}.
\newblock \showarticletitle{Pegasus: Pre-training with extracted gap-sentences
  for abstractive summarization}. In \bibinfo{booktitle}{\emph{International
  Conference on Machine Learning}}. PMLR, \bibinfo{pages}{11328--11339}.
\newblock


\bibitem[\protect\citeauthoryear{Zhang and Bansal}{Zhang and Bansal}{2019}]%
        {zhang2019addressing}
\bibfield{author}{\bibinfo{person}{Shiyue Zhang} {and} \bibinfo{person}{Mohit
  Bansal}.} \bibinfo{year}{2019}\natexlab{}.
\newblock \showarticletitle{Addressing semantic drift in question generation
  for semi-supervised question answering}.
\newblock \bibinfo{journal}{\emph{arXiv preprint arXiv:1909.06356}}
  (\bibinfo{year}{2019}).
\newblock


\bibitem[\protect\citeauthoryear{Zhang, Yin, Chen, Hung, Huang, and Cui}{Zhang
  et~al\mbox{.}}{2020a}]%
        {zhang2020gcn}
\bibfield{author}{\bibinfo{person}{Shijie Zhang}, \bibinfo{person}{Hongzhi
  Yin}, \bibinfo{person}{Tong Chen}, \bibinfo{person}{Quoc Viet~Nguyen Hung},
  \bibinfo{person}{Zi Huang}, {and} \bibinfo{person}{Lizhen Cui}.}
  \bibinfo{year}{2020}\natexlab{a}.
\newblock \showarticletitle{Gcn-based user representation learning for unifying
  robust recommendation and fraudster detection}. In
  \bibinfo{booktitle}{\emph{Proceedings of the 43rd International ACM SIGIR
  Conference on Research and Development in Information Retrieval}}.
  \bibinfo{pages}{689--698}.
\newblock


\bibitem[\protect\citeauthoryear{Zhao, Xiao, Gan, Zhang, and Xia}{Zhao
  et~al\mbox{.}}{2020}]%
        {zhao2020maintaining}
\bibfield{author}{\bibinfo{person}{Bowen Zhao}, \bibinfo{person}{Xi Xiao},
  \bibinfo{person}{Guojun Gan}, \bibinfo{person}{Bin Zhang}, {and}
  \bibinfo{person}{Shu-Tao Xia}.} \bibinfo{year}{2020}\natexlab{}.
\newblock \showarticletitle{Maintaining discrimination and fairness in class
  incremental learning}. In \bibinfo{booktitle}{\emph{Proceedings of the
  IEEE/CVF Conference on Computer Vision and Pattern Recognition}}.
  \bibinfo{pages}{13208--13217}.
\newblock


\bibitem[\protect\citeauthoryear{Zhao, Ni, Ding, and Ke}{Zhao
  et~al\mbox{.}}{2018}]%
        {zhao2018paragraph}
\bibfield{author}{\bibinfo{person}{Yao Zhao}, \bibinfo{person}{Xiaochuan Ni},
  \bibinfo{person}{Yuanyuan Ding}, {and} \bibinfo{person}{Qifa Ke}.}
  \bibinfo{year}{2018}\natexlab{}.
\newblock \showarticletitle{Paragraph-level neural question generation with
  maxout pointer and gated self-attention networks}. In
  \bibinfo{booktitle}{\emph{Proceedings of the 2018 Conference on Empirical
  Methods in Natural Language Processing}}. \bibinfo{pages}{3901--3910}.
\newblock


\bibitem[\protect\citeauthoryear{Zhou, Yang, Wei, Tan, Bao, and Zhou}{Zhou
  et~al\mbox{.}}{2017}]%
        {zhou2017neural}
\bibfield{author}{\bibinfo{person}{Qingyu Zhou}, \bibinfo{person}{Nan Yang},
  \bibinfo{person}{Furu Wei}, \bibinfo{person}{Chuanqi Tan},
  \bibinfo{person}{Hangbo Bao}, {and} \bibinfo{person}{Ming Zhou}.}
  \bibinfo{year}{2017}\natexlab{}.
\newblock \showarticletitle{Neural question generation from text: A preliminary
  study}. In \bibinfo{booktitle}{\emph{National CCF Conference on Natural
  Language Processing and Chinese Computing}}. Springer,
  \bibinfo{pages}{662--671}.
\newblock


\bibitem[\protect\citeauthoryear{Zhou, Zhang, and Wu}{Zhou
  et~al\mbox{.}}{2019}]%
        {zhou2019question}
\bibfield{author}{\bibinfo{person}{Wenjie Zhou}, \bibinfo{person}{Minghua
  Zhang}, {and} \bibinfo{person}{Yunfang Wu}.} \bibinfo{year}{2019}\natexlab{}.
\newblock \showarticletitle{Question-type driven question generation}.
\newblock \bibinfo{journal}{\emph{arXiv preprint arXiv:1909.00140}}
  (\bibinfo{year}{2019}).
\newblock


\bibitem[\protect\citeauthoryear{Zhu, Zhou, and Xia}{Zhu et~al\mbox{.}}{2020}]%
        {zhu2020generating}
\bibfield{author}{\bibinfo{person}{Yi Zhu}, \bibinfo{person}{Yiwei Zhou}, {and}
  \bibinfo{person}{Menglin Xia}.} \bibinfo{year}{2020}\natexlab{}.
\newblock \showarticletitle{Generating Semantically Valid Adversarial Questions
  for TableQA}.
\newblock \bibinfo{journal}{\emph{arXiv preprint arXiv:2005.12696}}
  (\bibinfo{year}{2020}).
\newblock


\end{thebibliography}

\appendix

\section{Implementation Details}
\subsection{Dataset Details}\label{sec:data_details}
\begin{table*}[]
  \centering
  \caption{Statistics of training/dev/testing sets for all QG datasets used in this paper.}
  \begin{tabular}{l|ccc}
  \hline
  \textbf{Dataset} & \textbf{Splitting Size}    & \textbf{Input Length avg.} & \textbf{Output Length avg.} \\ \hline
  SQuAD            & 87599/5286/5285  & 126.2/125.6/129.9          & 10.1/10.1/10.2              \\
  NarrativeQA      & 65494/6922/21114 & 578.4/558.2/574.3          & 8.6/8.4/8.7                 \\
  RACE             & 87866/4887/4930  & 314.7/311.5/314.4          & 11.5/11.4/11.5              \\
  McTest           & 1480/160/160     & 242.6/235.5/247.8          & 7.8/7.8/7.9                 \\
  OpenbookQA       & 4955/500/500     & 176.1/179.1/174.4          & 10.7/10.3/10.3              \\
  Arc-easy         & 2251/570/2376    & 214.3/212.1/214.4          & 19.1/19.7/19.6              \\
  Arc-hard         & 1119/299/1172    & 215.9/217.3/218.1          & 21.8/22.9/22.6              \\
  BoolQA           & 9427/1635/1635   & 99.4/98.5/98.3             & 8.8/8.7/8.7                 \\ \hline
  \end{tabular}
  \label{tb_dataset_statistics}
  \end{table*}
Table.~\ref{tb_dataset_statistics} presents the statistics of the above datasets, including data splitting size, the average input length (our unified QG encoding's input length), and the average length of output (questions).

SQuAD is one of the most commonly used answer-extraction QG datasets. There are two versions, and we use SQuADv1.1~\cite{rajpurkar2016squad} in our experiment, which contains $536$ Wikipedia articles with more than $100$k questions.
Following~\cite{yuan2021improving}, we split the dataset into training, dev, and testing sets, each set containing $87,599$, $5,286$, and $5,285$ elements.
NarrativeQA~\cite{kovcisky2018narrativeqa} is a collection of about $1,500$ stories and movie scripts with summaries. About $47,000$ question and answer pairs are created by crowd-source workers.
The dataset is splitted into training, dev, and testing sets with the size of $65,494$, $6,922$, and $21,114$, respectively.

RACE, McTest, OpenbookQA, ARC-easy, and ARC-hard are multi-choice datasets.
RACE~\cite{lai2017race} collects nearly $98,000$ multi-choice questions on $28,000$ articles from middle and high school English tests.
Since tests are constructed by experts, the questions are more challenging and often rely on multiple sentences.
In this paper, we divide RACE into $87,866$, $4,887$, and $4,930$ sets for training, dev, and testing.
McTest~\cite{richardson2013mctest} comprises of about $500$ fictional stories and $2,000$ questions formed in multi-choice.
Most of the questions are open-domain with less restriction on what can be asked.
The size of training, dev, and testing set for McTest are $1,480$, $160$, and $160$, respectively.
OpenbookQA~\cite{mihaylov2018can} contains about $6,000$ open-domain science questions and it provides an "open-book" of about $1,300$ science facts.
Since the question is associated with some commonsense knowledge, generating such questions are more challenging.
We split the dataset into training, dev and testing sets with sizes 4,955, 500 and 500 respectively.
Arc-easy and Arc-hard~\cite{clark2018think} questions are derived from ARC multi-choice question set released as part of the AI2 Reasoning Challenge in 2018\footnote{https://allenai.org/data/arc}.
ARC provides about $8,000$ questions as well as $14$ million sentences about science that are related to the questions.
For Arc-easy, the training, dev, and testing sets contain $2,251$, $570$, and $2,376$ examples respectively.
And for Arc-hard, there are $1,119$, $299$, and $1,172$ examples in the training, dev and testing sets respectively.

BoolQA~\cite{clark2019boolq} contains $15,942$ yes/no questions gathered from anonymized and aggregated queries to the Google search engine.
The queries are processed by using the pipeline from NQ~\cite{kwiatkowski2019natural} combined with an additional filtering step to focus on yes/no questions.
In this paper, we split the dataset into training, dev, and testing sets with the size of $9,427$, $1,635$, and $1,635$, respectively.

\subsection{Model Implementation}\label{sec_imp_details}
  All of our approaches are implemented based on PyTorch\footnote{https://pytorch.org/}.
  The backbone of our QG model is T5.
  Since T5 has a wide range of sizes version, we choose T5-base as the initial point of our QG model.

  We utilize AdamW~\cite{loshchilov2017decoupled} as our optimizer with $3e-4$ learning rate.
  Gradient clipping is applied during the training period.
  For each dataset, we train at most $20$ epochs, and if the validation loss does not decrease after $3$ epochs, the training process will be early stopped.
  The batch size is $16$.
  The max length of the input is set to $512$.
  The original $\lambda_{ori}$ for EWC term is set to $120000$ in practice.

  \section{Improving QA systems using Unified-QG} \label{sec_qa_qg}
One of the important applications of QG is to improve QA system's performance via training data augmentation.
Traditional QG models can only produce synthetic QA pairs for specific QA tasks.
Therefore, in order to improve different kinds of QA systems, we have to construct, train, and store a large amount of QG models. 
As a result, the computational costs and storage space of the QG model are equal to or even greater than QA models'.
In this part, we investigate how to use our \emph{single} Unified-QG to improve $8$ kinds of QA systems' performance, showing the efficiency of our Unified-QG.

To be simplified and without loss of generality, we use T5 as the backbone of our QA model and construct a unified QA encoding to let the model has the ability to address different QA tasks, instead of using dedicated QA models.
Similar to our unified QG encoding, the input for QA model is also a string concatenation: ``Question: ... + Passage: ... + [Distractor: ... +[...]]'', and the output target is the answer.
Then, we train this QA model on different QA datasets respectively to investigate the benefits of QG augmentation.

To generate synthetic QA pairs, we follow the ``back translation'' approach in ~\cite{sennrich2015improving,zhang2019addressing}. 
We utilize our Unified-QG model to generate questions based on the context in original dataset, without introducing new articles.
The low quality synthetic data are filtered out by using BLEU-4 scores.
For all of the QA datasets, the size of the augmented dataset is two times of original one.
Finally, we train QA model on the augmented dataset.

Following Ding et al's work~\cite{dong2019unified}, we evaluate each dataset with the metrics most frequently used in previous works.
Specifically, we employ exact match (EM) and ROUGE-L for extraction and abstraction QA.
The boolean QA is evaluated by the accuracy of the ``yes'' or ``no'' label.
The measure of multi-choice QA is relatively complex, since T5 directly generates the choice.
We at first use word overlap count to find the closest choice for the generated answer, and then, we calculate how often it is correct.

  \begin{table}[!h]
    \renewcommand{\arraystretch}{0.9}
    \centering
    \caption{Improving QA tasks using Unified-QG generated data.}
    \begin{tabular}{lccc}
    \hline
    \textbf{Dataset}     & \textbf{Metrics} & \textbf{original} & \textbf{combined} \\ \hline
    \textbf{SQUAD}       & Exact Mach       & 80.03             & \textbf{80.38}    \\
    \textbf{NarrativeQA} & ROUGE-L          & 25.23             & \textbf{25.63}    \\
    \textbf{RACE}        & Accuracy         & 57.91             & \textbf{58.48}    \\
    \textbf{McTest}      & Accuracy         & 65.62             & \textbf{70.63}    \\
    \textbf{OpenbookQA}  & Accuracy         & 51.60             & \textbf{53.40}    \\
    \textbf{Arc-easy}    & Accuracy         & 74.78             & \textbf{74.91}    \\
    \textbf{Arc-hard}    & Accuracy         & 34.47             & \textbf{34.56}    \\
    \textbf{BoolQA}      & Accuracy         & 75.41             & \textbf{77.37}    \\ \hline
    \end{tabular}
    \label{tb_improve_qa}
    \end{table}

Table~\ref{tb_improve_qa} presents the results of improving $8$ QA tasks via using Unified-QG generated data.
From Table~\ref{tb_improve_qa}, we can observe that all the QA systems are improved after trained on combined datasets.
To be specific, QA gains $0.35$ EM scores on SQuAD, $0.40$ ROUGE-L scores on NarrativeQA, $0.57$ accuracy scores on RACE, $5.01$ accuracy scores on McTest, $1.80$ scores on OpenbookQA, $0.13$ accuracy on Arc-easy, $0.09$ scores on Arc-hard, and $1.96$ accuracy on BoolQA.
It is worth mentioning that since the aim of this experiment is to demonstrate that our single Unified-QG can simultaneously improve multiple QA tasks, which is infeasible for traditional QG models, therefore, the data filter method and synthetic data usages here are relatively simple.
The advanced usage of QG for QA can be further studied in the future research.

\end{document}